\documentclass[twocolumn,twoside,10pt,final]{article}
\pdfoutput=1
\usepackage{framed,multirow}
\usepackage{fancyhdr}
\usepackage{amssymb}
\usepackage{latexsym}

\usepackage{amsmath}
\usepackage{xurl}
\usepackage{xcolor}

\usepackage{hyperref}
\usepackage{gensymb}
\usepackage[subrefformat=parens, labelformat=parens]{subcaption}
\usepackage{textcomp}
\usepackage{booktabs}
\usepackage[export]{adjustbox}
\usepackage[stable]{footmisc}

\usepackage{graphicx}
\graphicspath{{Figs/}}
\usepackage{natbib}

\usepackage[font=scriptsize]{caption}

\begin{document}
\twocolumn[
  \begin{@twocolumnfalse}
  \thispagestyle{empty}
\title{SERV-CT: A disparity dataset from CT for validation of endoscopic 3D reconstruction}
\date{}

\vspace*{-50pt}
\begin{minipage}{\textwidth}
\centering
\author{
\textsuperscript{*}P.J. ``Eddie'' Edwards\textsuperscript{ a},
\textsuperscript{*}Dimitris Psychogyios\textsuperscript{ a}
Stefanie Speidel\textsuperscript{ b} \\
Lena Maier-Hein\textsuperscript{ c} 
Danail Stoyanov\textsuperscript{ a}
}
\end{minipage}

\maketitle

\vspace*{-20pt}\hspace*{40pt}\begin{minipage}{0.9\textwidth}
\footnotesize
\textsuperscript{a} \textit{Wellcome/EPSRC Centre for Interventional and Surgical Sciences (WEISS), UCL, Charles Bell House, 43-45 Foley Street, London W1W 7TS, UK} \newline
\textsuperscript{b} \textit{Division of Translational Surgical Oncology, National Center for Tumor Diseases (NCT) Dresden, 01307, Dresden, Germany} \newline
\textsuperscript{b} \textit{Division of Medical and Biological Informatics, German Cancer Research Center (DKFZ), Heidelberg, Germany}
\end{minipage}
\vspace*{20pt}
\begin{abstract}
In computer vision, reference datasets from simulation and real outdoor scenes have been highly successful in promoting algorithmic development in stereo reconstruction. Endoscopic stereo reconstruction for surgical scenes gives rise to specific problems, including the lack of clear corner features, highly specular surface properties and the presence of blood and smoke. These issues present difficulties for both stereo reconstruction itself and also for standardised dataset production. Publicly available datasets have been produced using CT and either phantom images or biological tissue samples with markers attached covering a relatively small region of the endoscope field-of-view. We present a stereo-endoscopic reconstruction validation dataset based on CT (SERV-CT). Two {\it ex vivo} small porcine full torso cadavers were placed within the view of the endoscope with both the endoscope and target anatomy visible in the CT scan. Subsequent orientation of the endoscope was manually aligned to match the stereoscopic view and reference, disparities and occlusions calculated. The requirement of a CT scan limited the number of stereo pairs to 8 from each {\it ex vivo} sample. For the second sample an RGB surface was acquired to aid alignment of smooth, featureless surfaces. Repeated manual alignments showed an RMS disparity accuracy of around 2 pixels and a depth accuracy of about 2mm. A simplified reference dataset is provided consisting of endoscope image pairs with corresponding calibration, disparities, depths and occlusions covering the majority of the endoscopic image and a range of tissue types, including smooth specular surfaces, as well as significant variation of depth. We assessed the performance of various stereo algorithms from online available repositories. There is a significant variation between algorithms, highlighting some of the challenges of surgical endoscopic images. The SERV-CT dataset provides an easy to use stereoscopic validation for surgical applications with smooth reference disparities and depths with coverage over the majority of the endoscopic images. This complements existing resources well and we hope will aid the development of surgical endoscopic anatomical reconstruction algorithms.
\end{abstract}
\vspace{20pt}\\

  \end{@twocolumnfalse}
]



\footnotetext{First and second authors have equal contribution}
\footnotetext{Corresponding author email: eddie.edwards@ucl.ac.uk}
\section{Introduction}
\label{sec:intro}

\begin{figure*}[!t]
\resizebox{1\textwidth}{!}{%
\includegraphics[width=1.0\textwidth, valign=c]{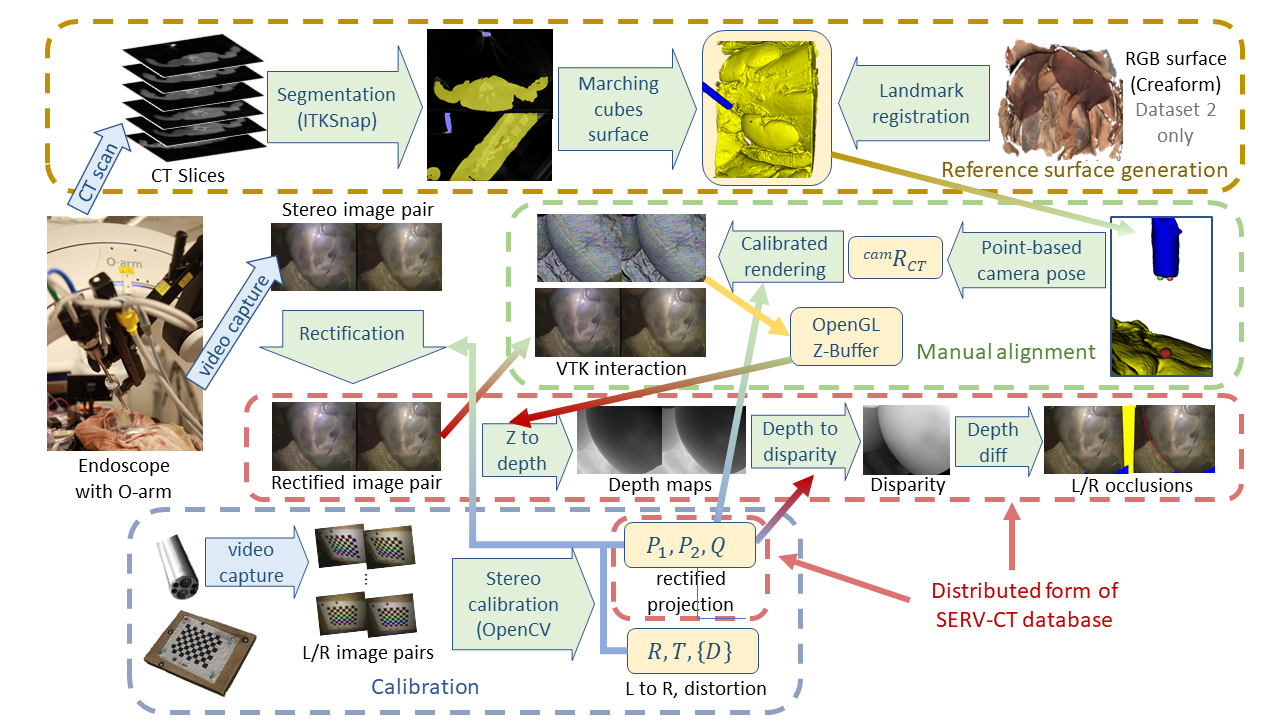}
}
\caption{A schematic flowchart showing the process of data generation, processing and alignment. A stereo image pair of an anatomical surface is captured along with a corresponding CT scan showing the location of both endoscope and anatomy, which can be modelled through segmentation. The position of the endoscope cameras is constrained to be near the end of the endoscope and the orientation is adjusted manually to match the stereo view. The depth from OpenGL is used to calculate the disparity. Stereo views, disparity and calibration are provided along with maps of occluded regions.}
\label{fig:Flowchart}
\end{figure*}

In minimally invasive surgery (MIS), endoscopic visualisation facilitates procedures performed through small incisions that have the potential advantages of lower blood loss and infection rates than open surgery as well as better cosmetic outcome for the patient. Despite the potential advantages of MIS, working within the limited endoscopic field-of-view (FoV) can make surgical tasks more demanding, which may lead to complications and adds significantly to the learning curve for such procedures. Knowing the position and orientation of the endoscope with respect to the 3D shape of the anatomy can facilitate surgical navigation and guidance, which has the potential to improve outcomes through techniques such as augmented reality (AR) to enhance perioperative surgical visualization and provide detailed, multi-modal anatomical information (\cite{bernhardt2017status, Edwards_2021}).
Vision has the potential to provide both the 3D shape of the surgical site and also the relative location of the camera within the 3D anatomy, especially when stereoscopic devices are used in robotic MIS (\cite{Mountney_2010, Stoyanov_2012}). A range of optical reconstruction approaches have been explored for endoscopy with computational stereo being by far the most popular due to the clinical availability of stereo endoscopes (\cite{MaierHein_2013, Rohl_2012, Chang_2013}). Despite recent major advances in computational stereo algorithms (\cite{Zhou_2020}), especially with deep learning models, in the surgical setting robust 3D reconstruction remains difficult due to various challenges including specular reflections and dynamic occlusions from smoke, blood and surgical tools.

While stereo endoscopy is routinely used in robotic MIS, the majority of endoscopic surgical procedures are performed using monocular cameras. With monocular endoscopes, 3D reconstruction can be approached as a non-rigid structure-from-motion (SfM) or simultaneous localisation and mapping (SLAM) problem (\cite{grasa2013visual}). This is typically more challenging than stereo because non-rigid effects and singularities need to be accounted for as the camera moves within the deformable surgical site. Alternative vision cues such as shading have also been explored historically but with limited success until recent promising results from monocular single image 3D using deep learning models (\cite{mahmoud2018live,Bae_2020}). Such approaches have been applied in general abdominal surgery (\cite{lin2016video}), sinus surgery (\cite{liu2018self}), bronchoscopy~(\cite{visentini2017deep}), and colonoscopy (\cite{mahmood2018deep, rau2019implicit}). While monocular reconstruction has been shown to be feasible and extremely promising, the accuracy and robustness of the surfaces produced still needs more development.

\begin{figure*}[!t]
\resizebox{1\textwidth}{!}{%
\begin{subfigure}{0.27\textwidth}
\includegraphics[width=1.0\textwidth, valign=c]{Endoscope_Oarm}
\subcaption{}
\label{fig:Equipment}
\end{subfigure}
\begin{subfigure}{0.73\textwidth}
\includegraphics[width=1.0\textwidth, valign=c]{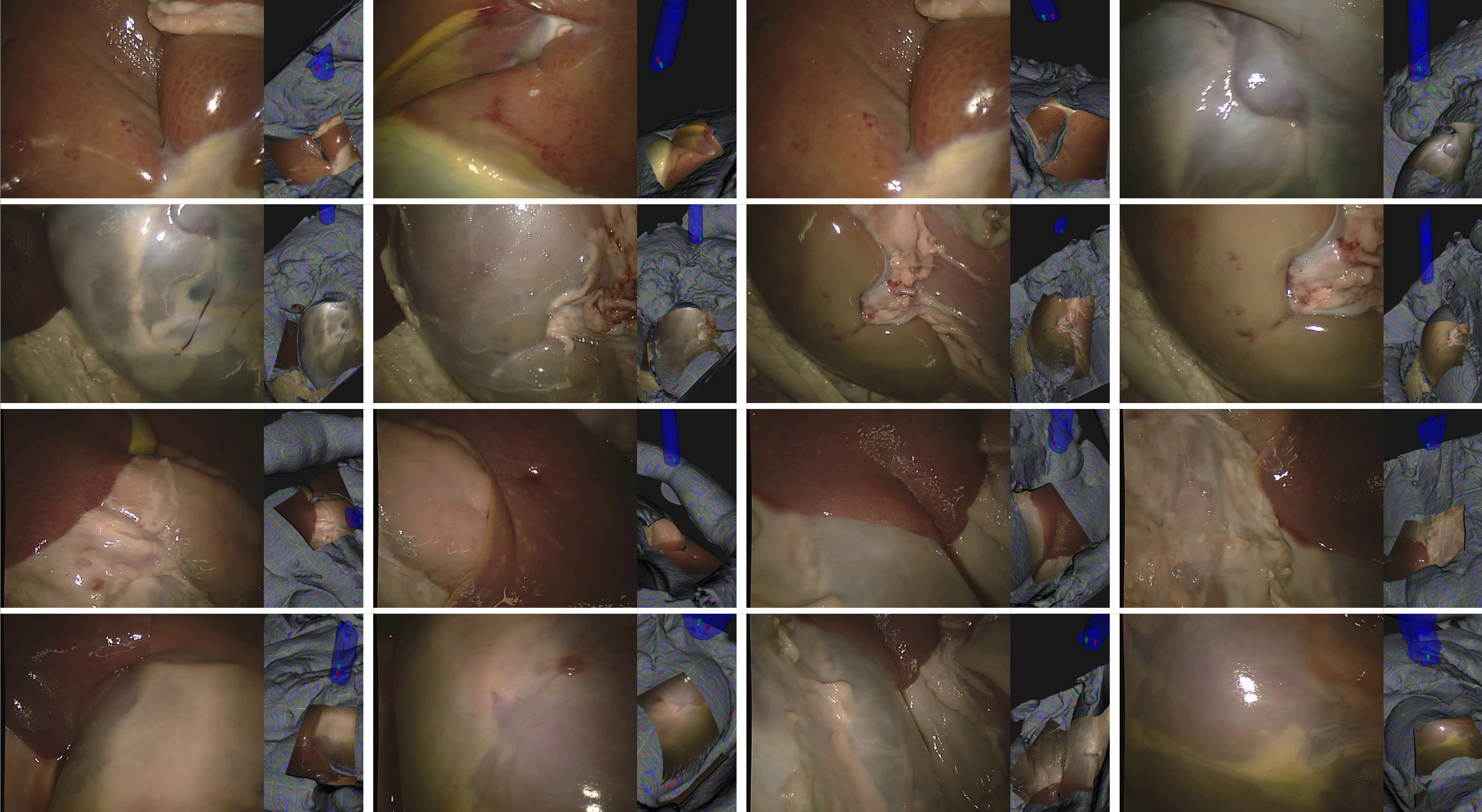}
\subcaption{}
\label{fig:AllImages}
\end{subfigure}
}
\caption{The equipment setup is shown, with endoscope attached to the da Vinci\texttrademark\ surgical robot placed within the O-arm\texttrademark\ interventional scanner~\subref{fig:Equipment}. The left images of all the views from the dataset alongside renderings of each endoscope view~\subref{fig:AllImages}. A range of tissue types is evident. The top row shows features with interesting variation of depth (image 1 and 3 are chosen to be similar to assess repeatability). The second row shows the kidney at different depths, with and without renal fascia. The third row shows the liver and surrounding tissues from the second {\it ex vivo} sample. The bottom row includes some smooth, featureless and highly specular regions from the kidney of sample 2. There is considerable variation of depth in most of these images (see Fig.~\ref{fig:disparity_depth_ranges}). }
\label{fig:EquipmentImages}
\end{figure*}

No matter the chosen method for 3D reconstruction, evaluation of the reconstruction accuracy and establishing appropriate benchmark datasets has been a major hurdle impeding development of the field. Standardised datasets providing accurate references have been instrumental for rapidly advancing the development of 3D reconstruction algorithms in computer vision (\cite{scharstein2003high, scharstein2014high, Menze2015CVPR}). In surgical applications, however, it has proved difficult to produce accurate standardised datasets in a form that facilitate easy and widespread use and adoption. Assessment of reconstruction accuracy requires 3D reference information and during in vivo surgical procedures this is not available. Phantom models made from synthetic materials have been used as surrogate environments with a corresponding gold standard from CT\footnote{\url{http://hamlyn.doc.ic.ac.uk/vision/}} (\cite{stoyanov2010real}). The CT model is registered to the stereo-endoscopic view using fiducials and dynamic CT is used to provide low frequency estimates of the phantom motion. More recently, the EndoAbS dataset was reported with stereo-endoscopic images of phantoms with gold standard depth provided by a laser rangefinder \cite{penza2018endoabs}\footnote{\url{https://zenodo.org/record/60593}}. Many challenging images are presented, including low light levels and smoke, and the dataset concentrates on the robustness of algorithms to these conditions. One of the main problems with phantom environments is their limited representation of the visual complexity of real {\it in vivo} images though some exciting progress in phantom fabrication and design, such as the work by \cite{Ghazi_2017}, may overcome this in the future.

Stereo-endoscopic datasets have also been created using {\it ex vivo} animal environments. The first contains endoscopic images of samples from different porcine organs (liver, kidney, heart) captured from various angles and distances including examples with smoke and artificial blood(~\cite{maier2014comparative}). Gold standard 3D reconstruction is provided within a masked region for each stereo pair from CT scans registered using markers visible in both the CT scan and the endoscopic images. Analysis tools are also provided within the MITK\footnote{\url{https://www.mitk.org}} framework (\cite{wolf2005medical}). The area of analysis is restricted to fairly small regions near the centre of the images leading to a comparatively narrow range of depths and tissue types within one sample. Most recently, as part of the Endoscopic Vision (EndoVis) series of challenges, the 2019 SCARED challenge\footnote{\url{https://endovissub2019-scared.grand-challenge.org/}} provides structured light surfaces reconstructed through the endoscope itself as a gold standard for stereo reconstruction. The HD endoscope images from {\it ex vivo} samples are brightly lit and the coverage of scene is excellent. Despite being very promising, the EndoVis data is so far provided for the purposes of the challenge only and has not yet been made freely available for general research.

A high quality validation dataset providing comprehensive image coverage of the endoscopic view for a range of tissue types and greater depth variation is needed to drive progress in 3D endoscopic reconstruction. In this paper, we provide such a dataset using a CT scan encompassing both the endoscope and the viewed anatomy. The aim was to use the CT endoscope position to avoid the need for CT markers visible in the endoscope view, allowing almost full coverage of the endoscope view. Example images of {\it ex vivo} porcine samples are provided that cover some of the more challenging and realistic examples of endoscopic views, including smooth featureless regions and specular surface properties. We present the methodology of constrained semi-manual alignment and provide both the dataset and algorithms in open source alongside all of the raw data\footnote{\url{https://www.ucl.ac.uk/interventional-surgical-sciences/serv-ct}}. In addition, we report results of some of the most recent open source computational stereo algorithms using this dataset as a baseline.

\section{The SERV-CT reference dataset construction}
\label{sec:gt_construction}

The working hypothesis of this research is that a CT scan containing both the endoscope and the viewed anatomical surface can be used to provide a sufficiently accurate reference for stereo reconstruction validation. To establish whether this is feasible, multiple stereo-endoscopic views of two {\it ex vivo} porcine full torsos were taken. The range of views can be seen in Fig.~\ref{fig:AllImages}. As a secondary aim we examine whether a textured RGB surface model can facilitate registration, particularly where there are few visible geometrical features to use for alignment.

\begin{figure}[!t]
\centering
\begin{subfigure}[b]{.45\linewidth}
\includegraphics[width=\linewidth]{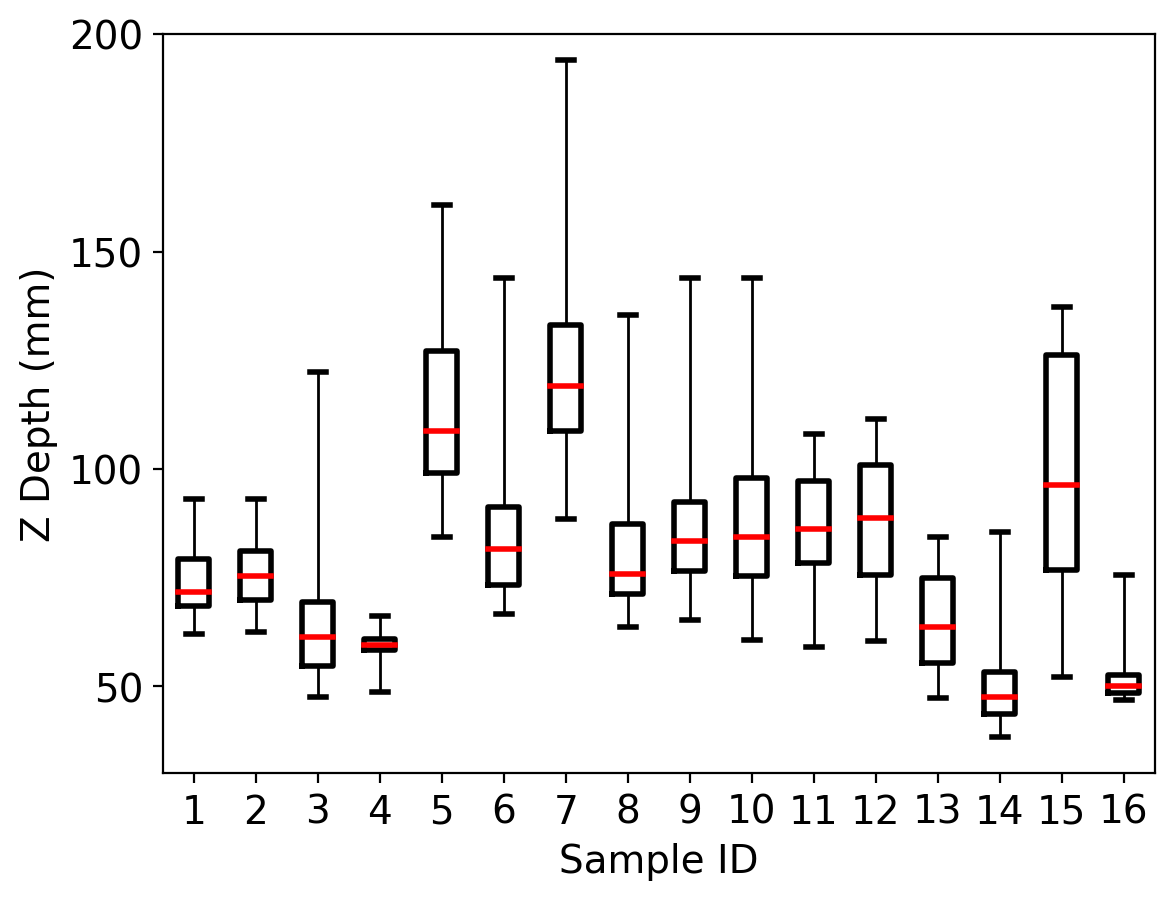}
\caption{Z-Depth Range}\label{fig:depth_range_ours}
\end{subfigure}
\begin{subfigure}[b]{.45\linewidth}
\includegraphics[width=\linewidth]{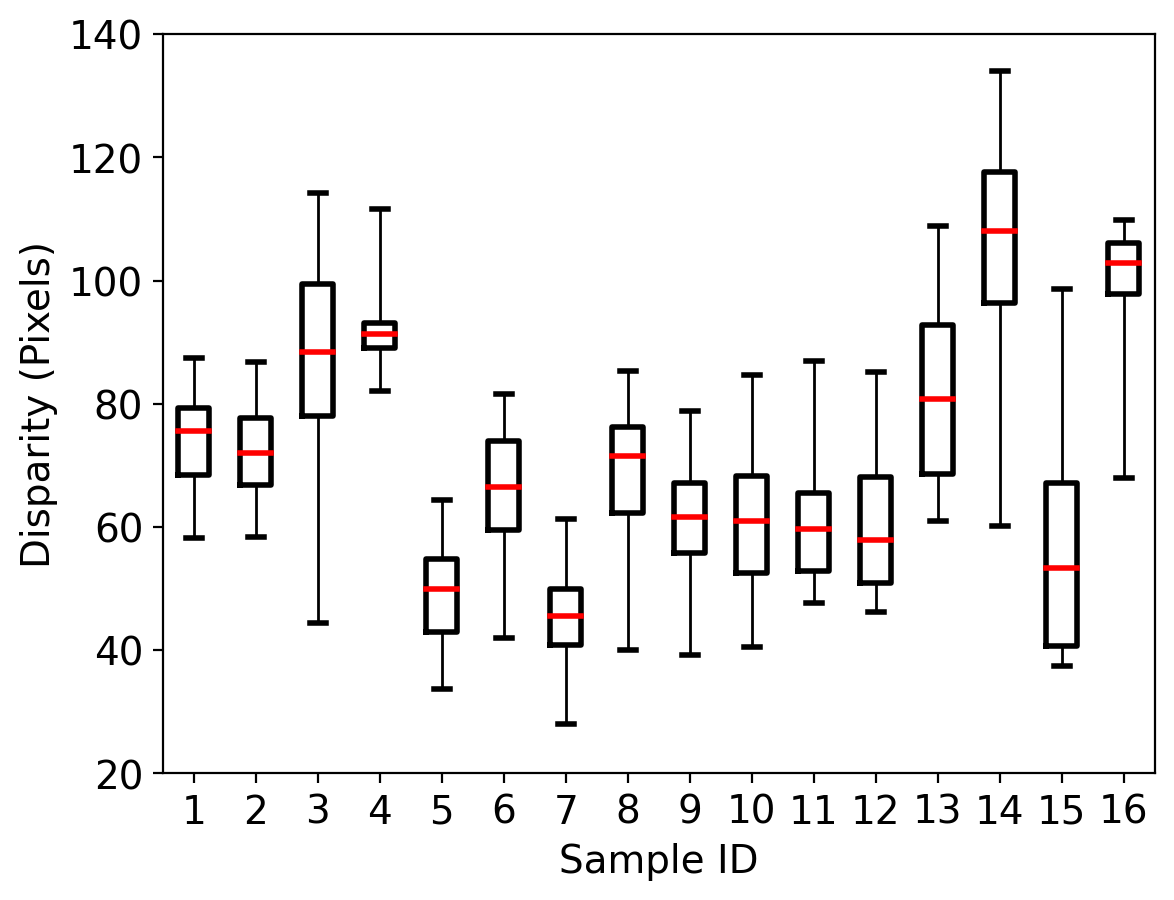}
\caption{Disparity Range}\label{fig:disparity_range_ours}
\end{subfigure}
\caption{Z-Depth and disparity ranges calculated from the reference standard for each sample in our dataset showing wide variation of depths provided by the sample images.}
\label{fig:disparity_depth_ranges}
\end{figure}

\subsection{Equipment}

As our anatomical model we used two fresh {\it ex vivo} full torso porcine cadavers including thorax and abdomen. In this study, we have focused on the abdomen as this is where most endoscopic surgery occurs. The outer tissue layers were removed by a clinical colleague in a manner that mimics surgical intervention to reveal the inner organs.

CT scans were provided using the O-arm\texttrademark\ Surgical Imaging System (Medtronic Inc., Dublin, Ireland). This interventional scanner provides 3-D scans equivalent to CT reconstructed from a rotating X-ray set enclosed within a circular casing. For the second dataset, in order to facilitate alignment in very smooth anatomical regions with few geometric features, the sample torso was also scanned with a Creaform Go SCAN 20 hand-held scanner (Creaform Inc., Lévis, Canada) to provide a  RGB structured light surface. 
 
Endoscope images are collected using a first generation da Vinci\texttrademark\ surgical system (Intuitive Surgical, Inc., Sunnyvale, CA, USA), which can also be used to manipulate the endoscope position. This robotic surgical setup is not ideally designed to fit within the confines of the O-arm\texttrademark\ and positioning requires considerable angulation of both the robot setup joints and the O-arm\texttrademark\ itself. The setup can be seen in Fig.~\ref{fig:Equipment}, showing the robotic endoscope within the O-arm\texttrademark\ and the {\it ex vivo} model. Images were gathered in two separate experiments using the straight and 30\degree\ endoscopes supplied with the da Vinci\texttrademark\ system.

\subsection{Endoscope calibration}
\label{sec:endo_cal}
Endoscope calibration follows a standard chessboard OpenCV\footnote{\url{https://opencv.org/}} stereo calibration protocol. Images of a chessboard calibration object are taken from multiple viewpoints the corners detected enable intrinsic calibration of the two cameras of the stereo endoscope. The same chessboard pattern viewed in corresponding left and right views can also provide the transformation from the left camera to the right camera. We used 18 images pairs in Expt. 1 and 14 image pairs in Expt. 2 covering the endoscope view and a range of depths.

Once the left and right images have been rectified, the projection for each eye is given by two matrices:
\begin{align*}
P_1 =  \left[  \begin{matrix}
   f & 0 & C_x^1 & 0 \\
   0 & f & C_y^1 & 0 \\
   0 & 0 & 1 & 0 \\
   \end{matrix} \right]
   ,
P_2 =  \left[ \begin{matrix}
   f & 0 & C_x^2 & T_x f \\
   0 & f & C_y^2 & 0 \\
   0 & 0 & 1 & 0
   \end{matrix} \right]
\end{align*}

Their elements combine to give a matrix, $Q$, that relates the disparity, $\delta$, at a pixel $(u, v)$ to a to 3D location $(x, y, z)$:
\begin{equation}
\label{eqn:Q_matrix}
k \left[ \begin{matrix} x \\ y \\ z \\ 1\end{matrix}\right] = Q \left[ \begin{matrix} u \\ v \\ \delta \\ 1\end{matrix} \right]
\end{equation}
where
\begin{equation*}
Q= \left[ \begin{matrix}
1 & 0 & 0 & -C_x^1\\
0 & 1 & 0 & -C_y^1\\
0 & 0 & 0 & f\\
1 & 0 & -1/T_x & (C_x^1 - C_x^2)
\end{matrix} \right]
\end{equation*}
For the simplified form of the released dataset, only these three matrices, $P_1$, $P_2$ and $Q$ are provided alongside the rectified left and right images with corresponding reference disparities, depths and occlusions are provided.

In addition to multiple views being used for camera calibration, several CT scans depicting the endoscope and the chessboard are taken. Six fixed spherical ceramic coloured markers can be seen in both the CT scan and the endoscope views (see Fig.~\ref{fig:ChessboardVerification}(a)). These can be used to validate our alignment method and provide accuracy measurement of the CT alignment process.

\subsection{CT segmentation of endoscope and anatomy}
\label{sec:CTseg}
\begin{figure}[!t]
\captionsetup[sub]{font=Large}
\resizebox{1\columnwidth}{!}{%
\begin{tabular}{cc}
\multicolumn{2}{c}{
\begin{subfigure}{\textwidth}
\includegraphics[width=1.0\linewidth]{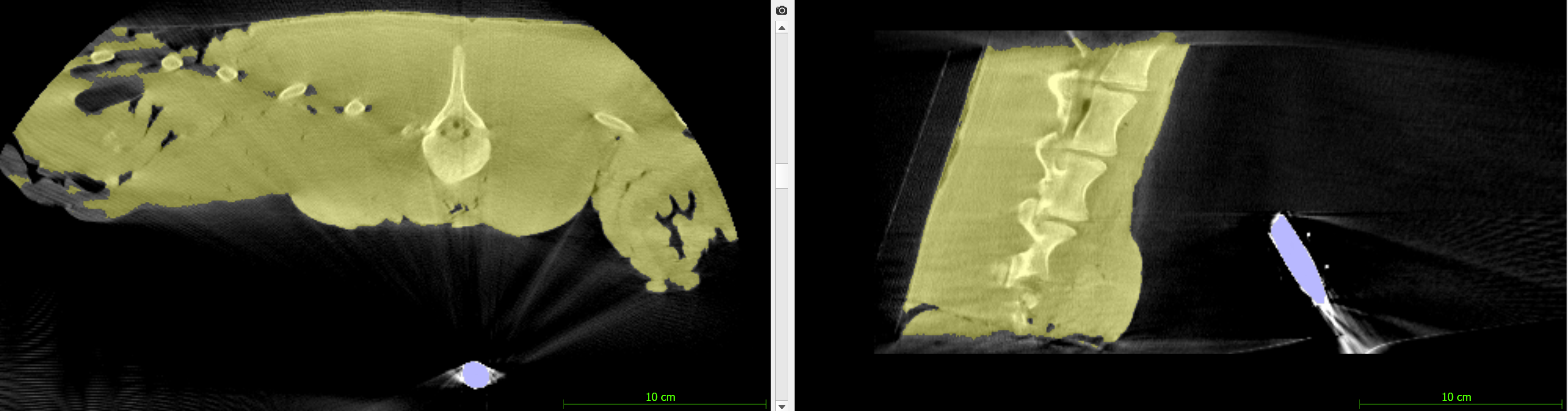}
\subcaption{}
\label{fig:CTseg}
\end{subfigure}
}
\\
\begin{subfigure}{0.5\textwidth}
\includegraphics[width=1.0\linewidth]{ExampleHandSegBefore}
\subcaption{}
\label{fig:CT_no_hand_seg}
\end{subfigure}
&
\begin{subfigure}{0.5\textwidth}
\includegraphics[width=1.0\linewidth]{ExampleHandSegAfter}
\subcaption{}
\label{fig:CT_hand_seg}
\end{subfigure}
\end{tabular}
}
\caption{O-arm\texttrademark\ CT scan showing our anatomical full torso model (yellow) and the endoscope (blue) visible in the same scan~\subref{fig:CTseg}. Streak artefacts from the presence of the metal endoscope are evident, but the viewed anatomical surface can still be accurately segmented. A single threshold does not always provide accurate segmentation~\subref{fig:CT_no_hand_seg} and these regions must be segmented by hand~\subref{fig:CT_hand_seg}}
\label{fig:CTsegmentation}
\end{figure}
As can be seen in Fig.~\ref{fig:CTseg}, both the endoscope and the anatomical surface are visible in the CT scan. The ITKSnap software (version 3.8.0) is used for its convenience and simple user interface (\cite{yushkevich2006user}). Automated segmentation with seeding and a single threshold is used initially. The endoscope is mostly at the fully saturated CT value so a very high threshold is used (2885). For the anatomical tissue-air interface, a very low value works well (-650 to -750). Partly due to the artifacts from the presence of the endoscope, but also because of thin membranes and air filled pockets, a single threshold does not capture all the anatomical surface. Some of the anatomical detail must be hand segmented (see Fig.~\ref{fig:CTsegmentation}). ITKSnap provides tools for this purpose, including pencils and adaptive brushes. Hand segmentation is limited to a few small regions, with most of the anatomical surface having clear contrast in the CT image.

\begin{figure}[!t]
\resizebox{1\columnwidth}{!}{%
\includegraphics[height=0.50\linewidth]{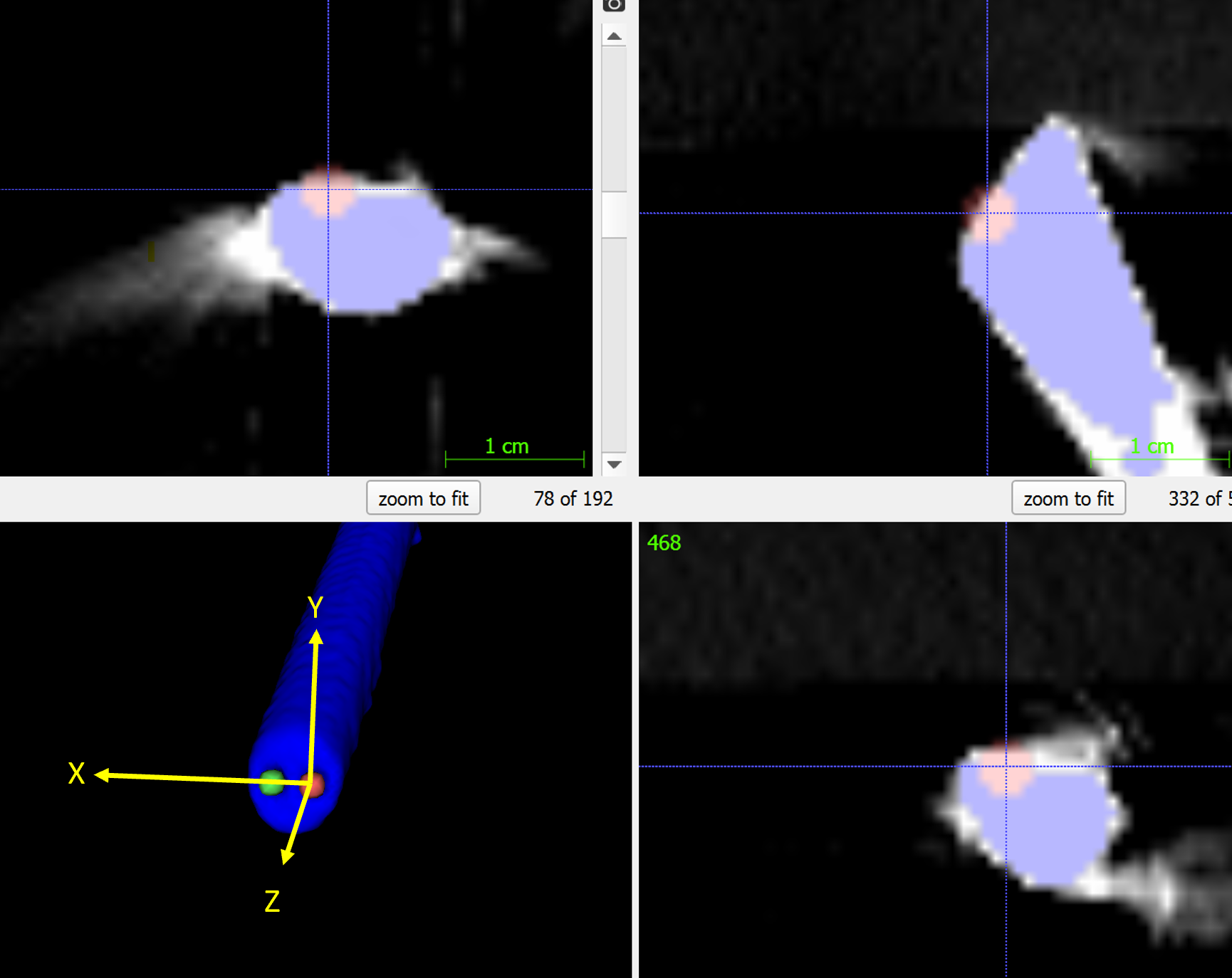}
\includegraphics[height=0.50\linewidth]{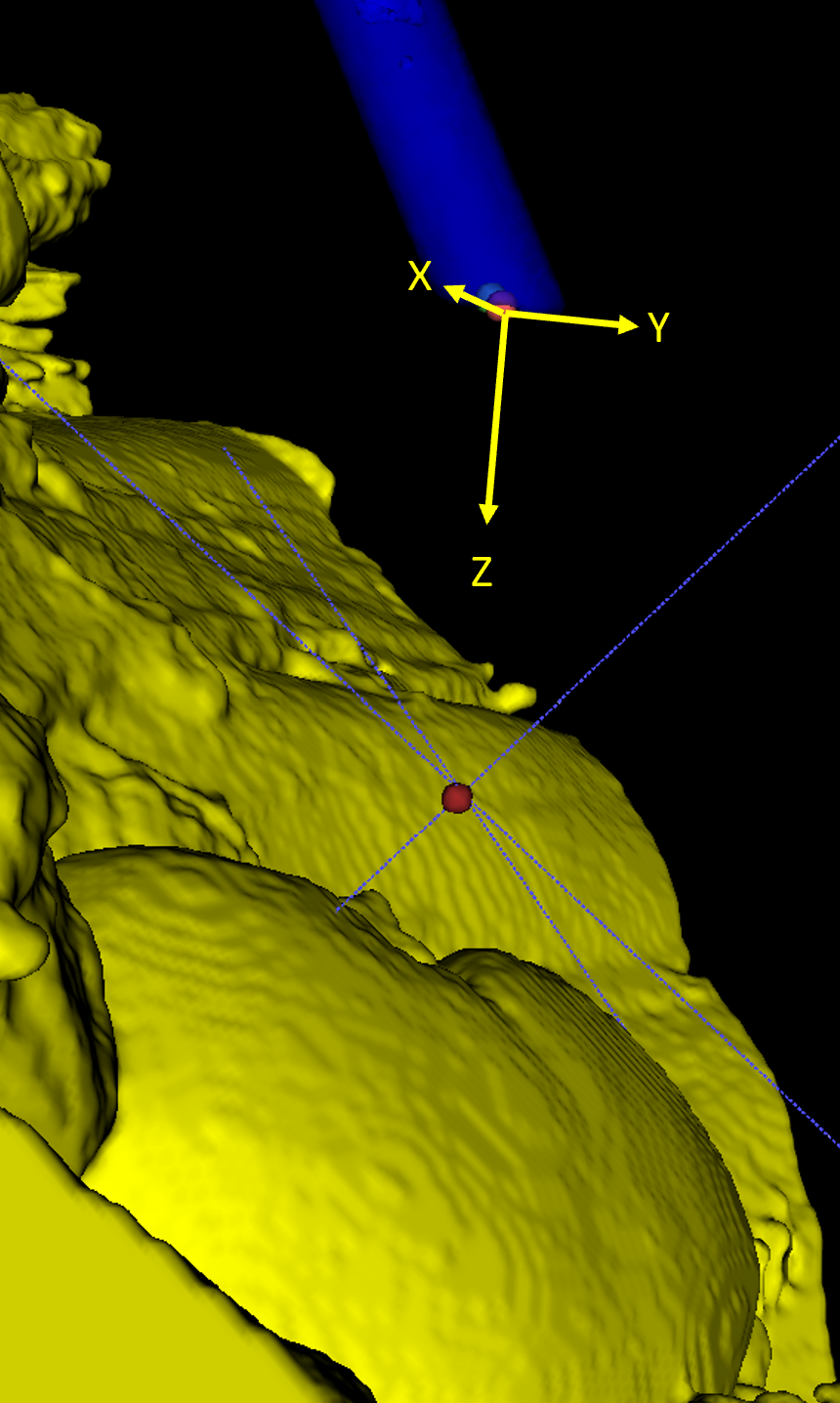}
}
\caption{Approximate left (red) and right (green) camera positions are marked on the endoscope surface in the CT scan. The left camera position is fixed in subsequent manual alignment as well as a target point roughly in the direction of the relevant anatomy. The resulting axes of this 30\degree endoscope are shown in yellow, with the Z axis pointing towards the target point. This provides an initial alignment from CT to endoscope, which can be subsequently refined manually.}
\label{fig:endo_orient}
\end{figure}

\subsection{Identification of CT endoscope position and orientation}
\label{sec:endoscope_orient}
To relate the coordinates of the endoscope to the CT, we first mark a rough position for the endoscope cameras. By rotating the surface rendered view of endoscope and anatomy, it is usually possible to identify which region of the CT is being viewed. Orienting the views accordingly, approximate positions for the left and right cameras can be marked on the end of the endoscope. It should be noted that the exact orientation is not important, as this will be adjusted manually later on. However, constraining the location of the camera to be near the end of the endoscope and largely limiting the motion to rotation about the left camera has two benefits. It reduces the complexity of the alignment task but more importantly it ensures that the distance from the camera to the anatomical surface is that described by the CT scan. Since stereoscopic disparity depends directly on this depth, a constrained motion preserving the camera position from the CT scan should lead to a more accurate reference. 

In addition to the left and right cameras, a further point along the endoscope is manually identified to provide an initial approximate viewing direction for the endoscope. The left and right camera positions and this point define an axis system and an initial estimate of the rigid transformation from CT to stereo endoscopic camera coordinates (see Fig.~\ref{fig:endo_orient}). The purpose of this initial rough alignment is partly to ensure that in the next phase of manual alignment, the relevant anatomy can be seen in a virtual rendering from the endoscope position. By constraining the endoscope camera position we ensure that the anatomy is viewed from the correct perspective, which is key to making an accurate reference.

\subsection{Verification using a calibration object}
\label{sec:calib_verify}

\begin{figure}
\captionsetup[sub]{font=Large}
\resizebox{1\columnwidth}{!}{%
\begin{tabular}{cccc}
\begin{subfigure}{0.29\textwidth}
\includegraphics[width=1.0\linewidth]{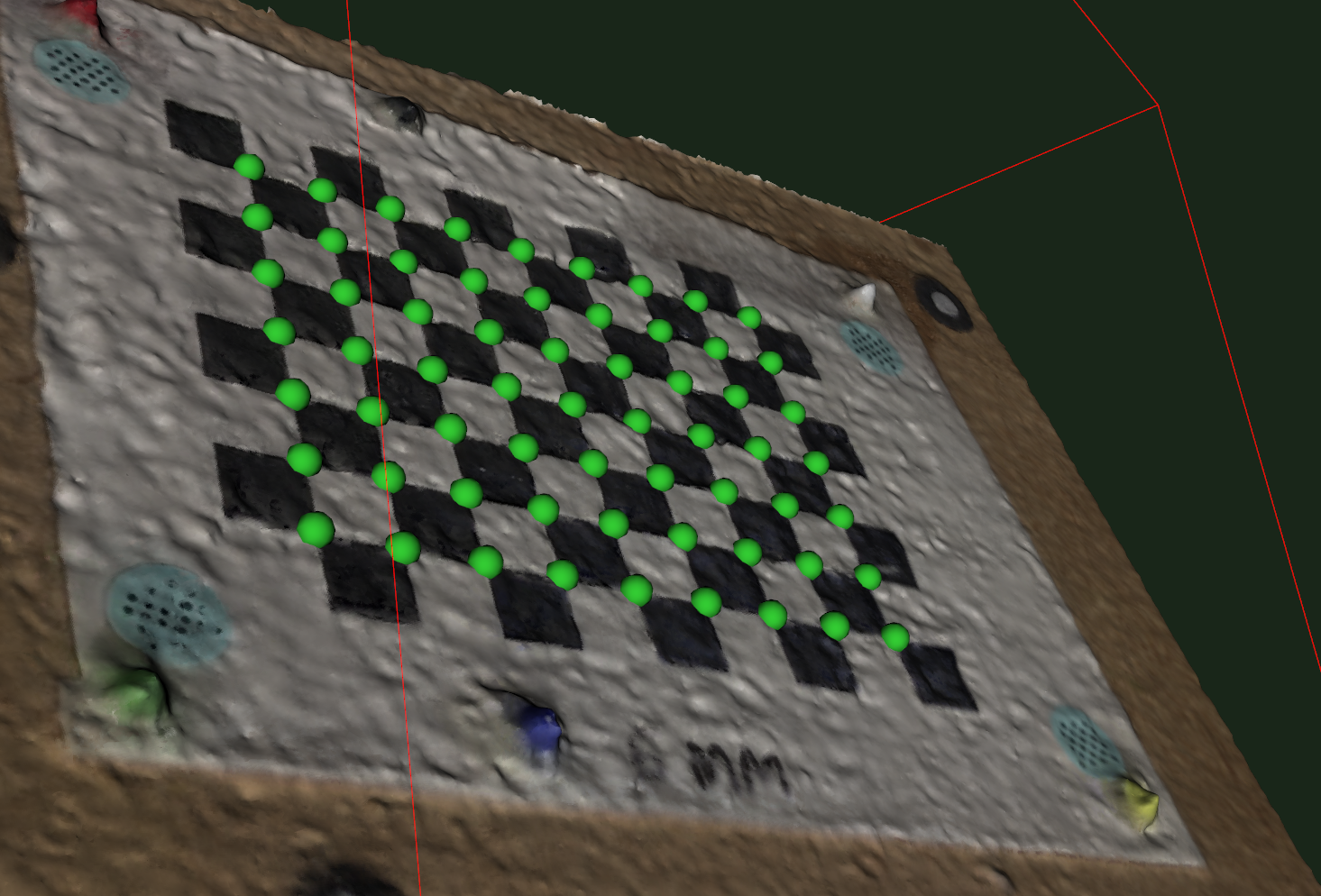}
\includegraphics[width=1.0\linewidth]{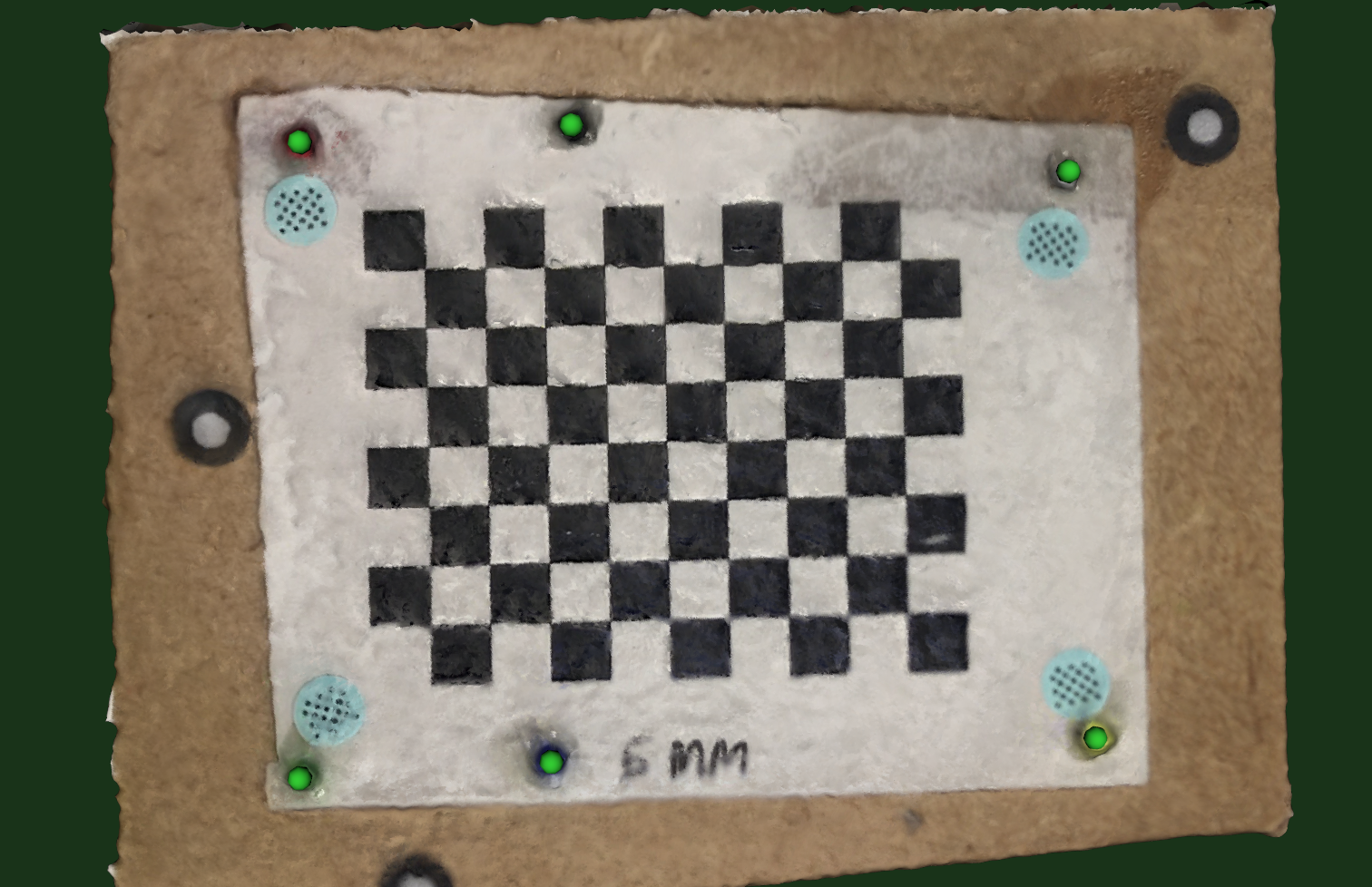}
\subcaption{}
\end{subfigure} &
\begin{subfigure}{0.24\textwidth}
\includegraphics[width=1.0\linewidth]{CalibNotRegToRecon}
\subcaption{}
\end{subfigure} &
\begin{subfigure}{0.26\textwidth}
\centering\includegraphics[width=0.87\linewidth]{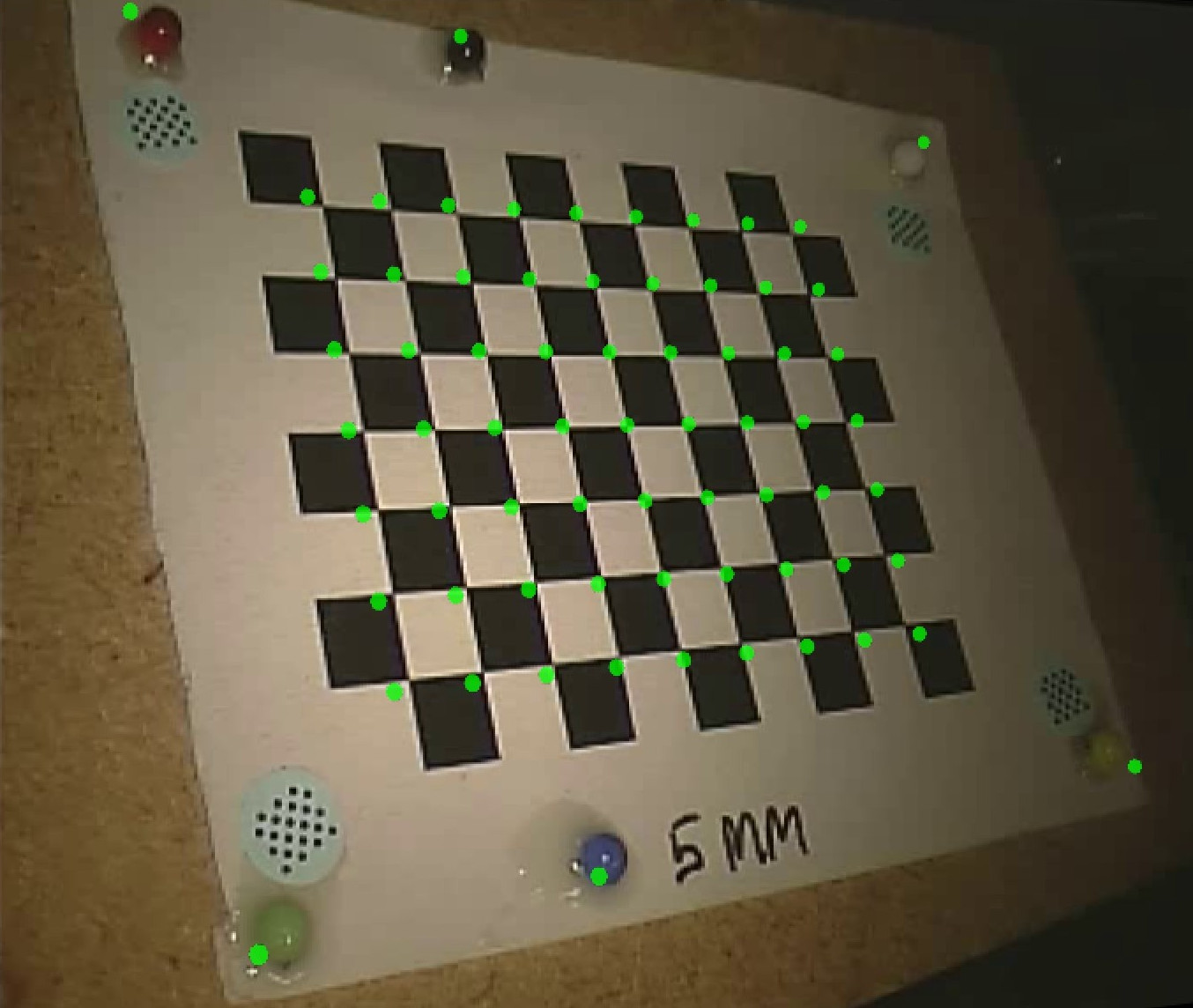}
\subcaption{}
\centering\includegraphics[width=0.87\linewidth]{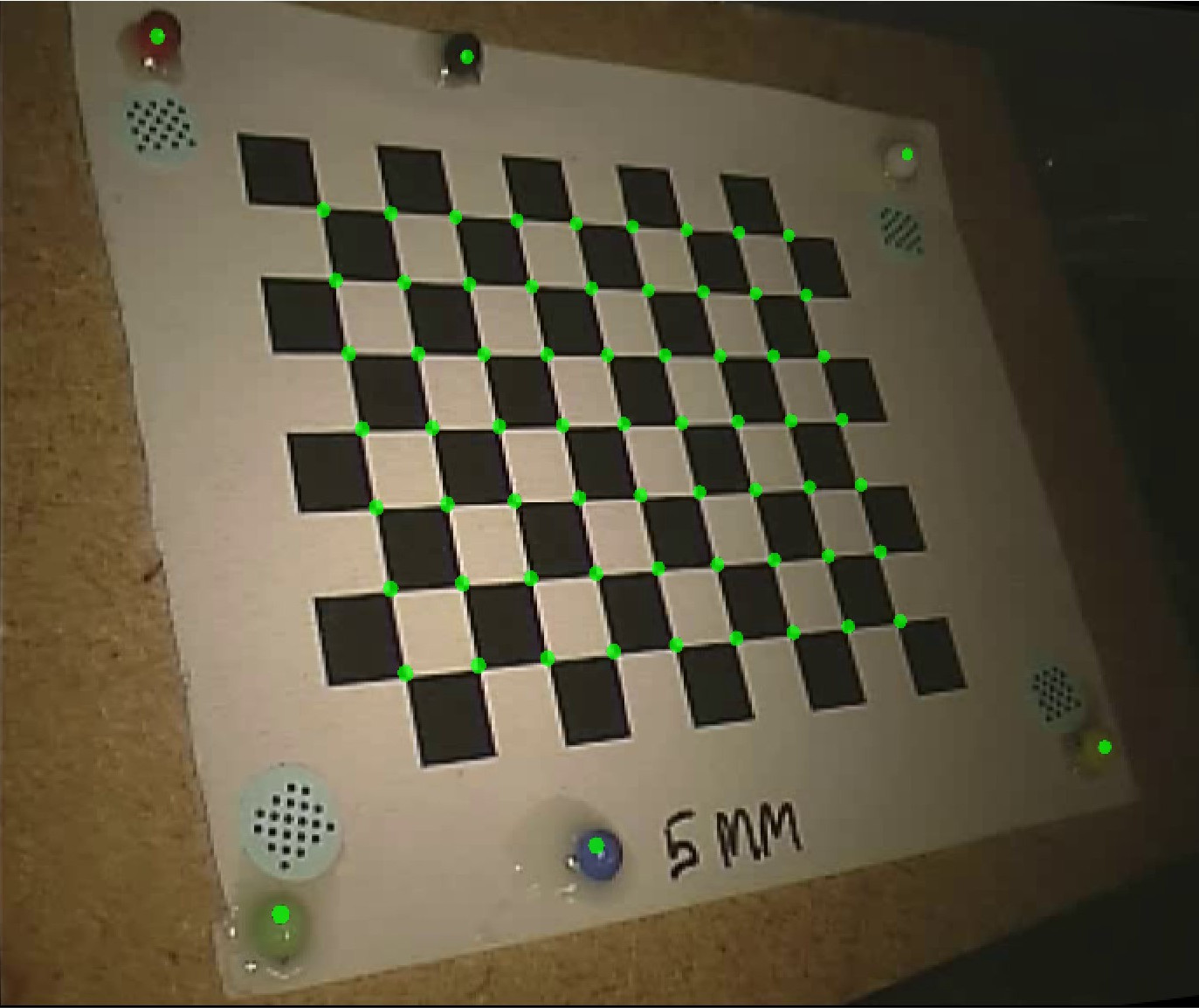}
\subcaption{}
\end{subfigure} &
\begin{subfigure}{0.24\textwidth}
\includegraphics[width=1.0\linewidth]{CalibRegToRecon}
\subcaption{}
\end{subfigure}
\end{tabular}
}
\caption{Meshroom reconstruction of the calibration object with chessboard points and fiducials marked (a). Rendering with the marked camera positions at the end of the endoscope (b). The scaling is slightly wrong (c). Translation of the cameras results in good alignment (d) and accurate reconstruction (e). This highlights the fact that we don't know the position of the effective pinhole of the endoscope cameras. Allowing a small translation into the body of the endoscope improves accuracy.}
\label{fig:ChessboardVerification}
\end{figure}

To establish an upper limit of accuracy for our method we have collected stereo images of the calibration object with a corresponding CT scan. Six coloured spherical bearings are used as fiducial markers. These can be readily identified in both the CT scan and the endoscopic view. To establish the location of the chessboard relative to the markers, a surface was produced from multiple images from an iPhone 7 using the {3D} reconstruction software, Meshroom\footnote{\url{https://github.com/alicevision/meshroom}}. Marking of both the fiducial markers and the chessboard points on this model provides a correctly scaled, aligned model and allows the fiducials to be expressed in chessboard coordinates (Fig.~\ref{fig:ChessboardVerification}(a)). Registration of the fiducial markers from the Meshroom model to the CT scan gives a residual alignment error of $<$0.5mm and the resulting chessboard points transformed to CT align well with the plane of the board.

Stereo reconstruction of the chessboard points (Fig.~\ref{fig:ChessboardVerification}(b)) exhibits a small scaling error (Fig.~\ref{fig:ChessboardVerification}(c)) which can be corrected for by moving the camera position back along the endoscope (Fig.~\ref{fig:ChessboardVerification}(d)). The initial camera positions are placed on the end of the endoscope surface, but the lens arrangement and optics of the endoscope is not known, so the effective pinholes may indeed be deeper inside the shaft. The resulting chessboard reconstruction is accurate ($\mathtt{\sim}$0.4mm) and the position of the cameras within the endoscope is reasonable (Fig.~\ref{fig:ChessboardVerification}(e)).

The process of manually adjusting the endoscope position and orientation to match the chessboard is the same as the alignment procedure discussed in the next section for registration of the anatomy, but the chessboard points give a measure of accuracy. Registration of the fiducial markers from the Meshroom model to the CT scan provides the chessboard points in CT coordinates, which align well with the segmented plane of the board (Fig.~\ref{fig:ChessboardVerification}(e)).

\subsection{Manual alignment of the endoscope orientation to match the anatomical surface}
\begin{figure*}
\captionsetup[sub]{font=normal}
\resizebox{1\textwidth}{!}{%
\begin{tabular}{cl}
\begin{tabular}{c}
\begin{subfigure}{0.65\textwidth}
\centering\includegraphics[width=1.0\linewidth]{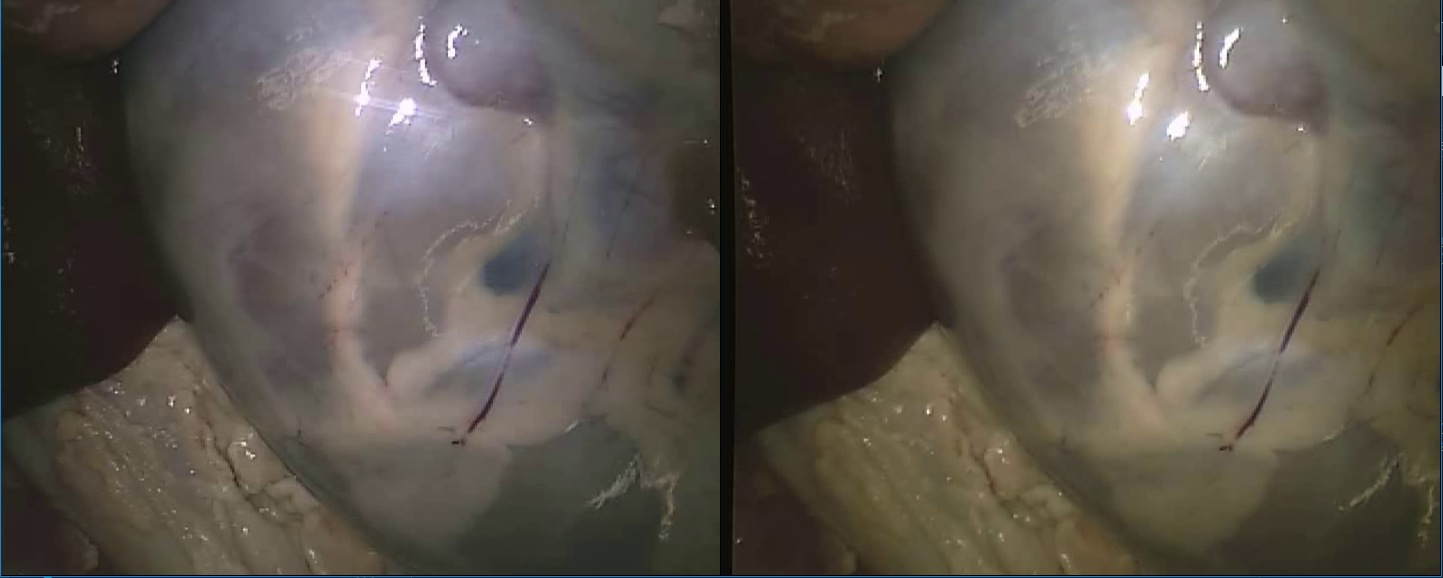}
\subcaption{}
\label{fig:StereoPair}
\end{subfigure}
\\
\begin{subfigure}{0.65\textwidth}
\centering\includegraphics[width=1.0\linewidth]{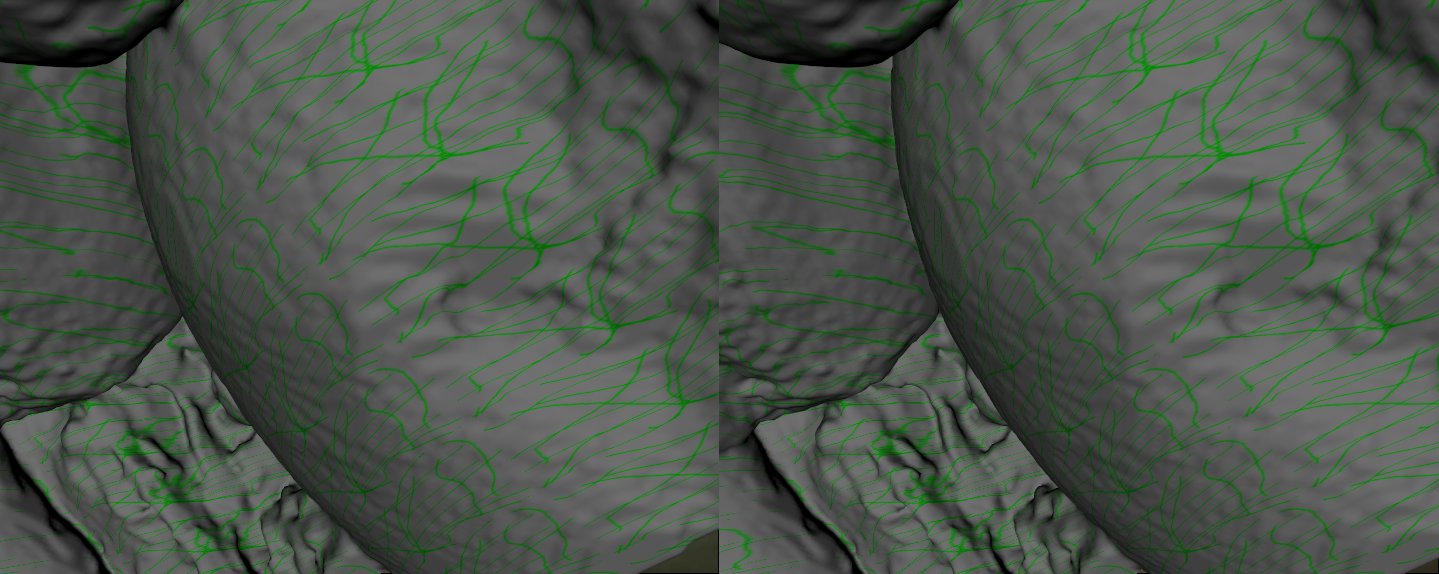}
\subcaption{}
\label{fig:StereoRendering}
\end{subfigure}
\\
\begin{subfigure}{0.65\textwidth}
\centering\includegraphics[width=1.0\linewidth]{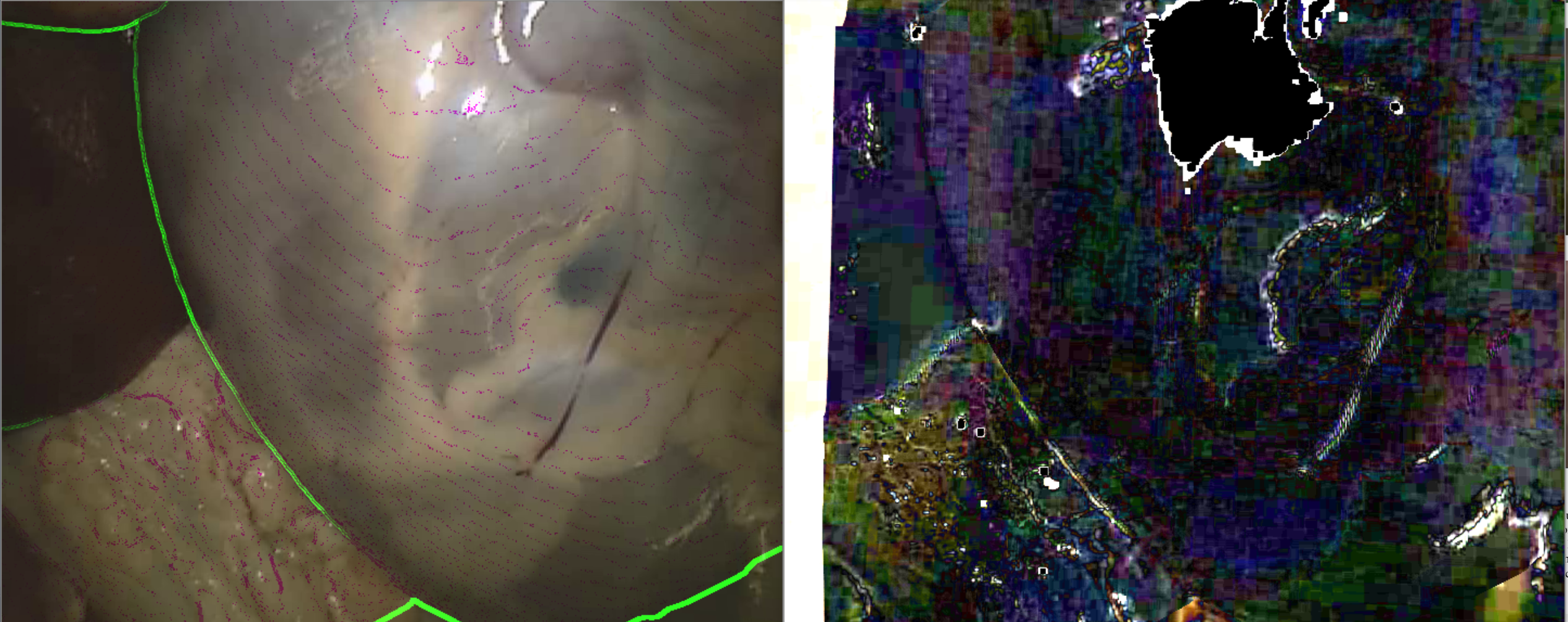}
\subcaption{}
\label{fig:HintsForAlignment}
\end{subfigure}
\end{tabular}
&
\begin{subfigure}{0.414\textwidth}
\centering\includegraphics[width=1.0\linewidth]{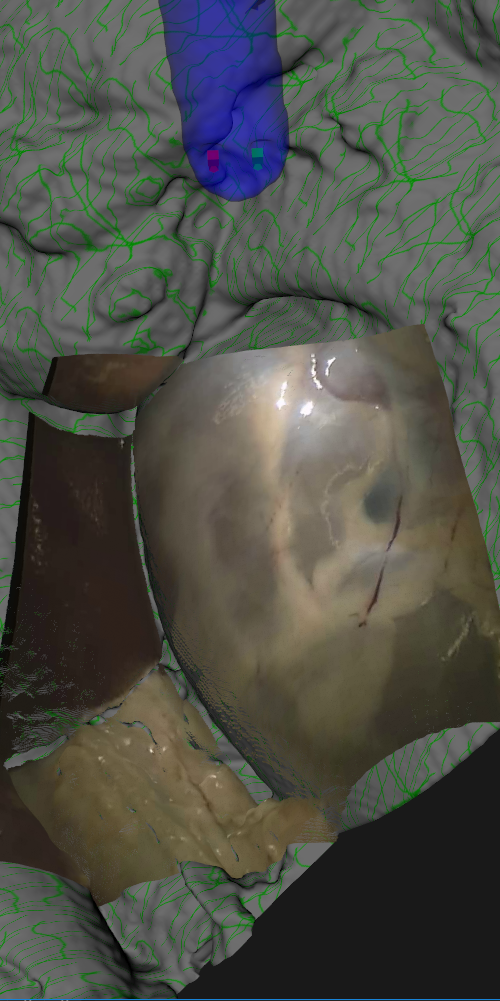}
\subcaption{}
\label{fig:StereoMappedToCT}
\end{subfigure}
\end{tabular}
}
\caption{The manual alignment method, showing the original image pair~\subref{fig:StereoPair}, the rendered image pair~\subref{fig:StereoRendering} and processed images to help with registration~\subref{fig:HintsForAlignment}. Overlays can be displayed using different renderings and manipulated to provide accurate alignment. Projection of the left image onto the CT surface after alignment is shown in (c). {\it In all cases in the paper the left image of a stereo pair is shown on the right to facilitate cross-eyed fusion}} 
\label{fig:ManualAlignment}       
\end{figure*}

\begin{figure*}
\resizebox{1\textwidth}{!}{%
\begin{tabular}{cl}
\begin{tabular}{c}
\begin{subfigure}{0.55\textwidth}
\centering\includegraphics[width=1.0\linewidth]{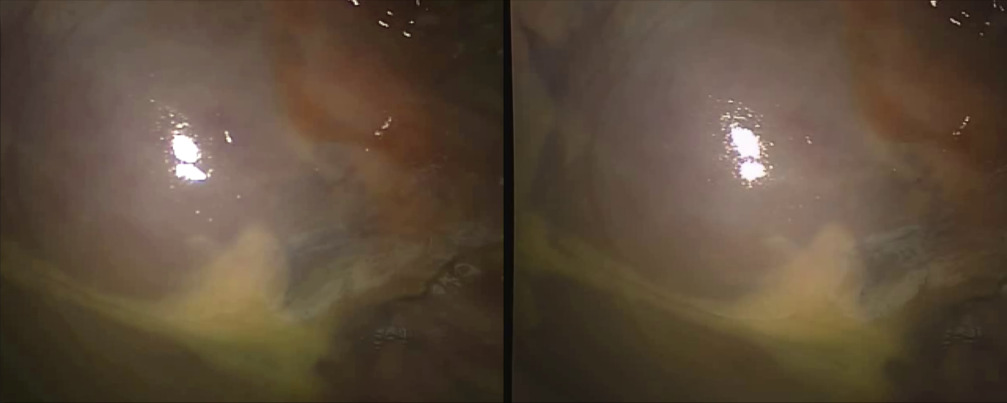}
\subcaption{}
\label{fig:SmoothStereoPair}
\end{subfigure}
\\
\begin{subfigure}{0.55\textwidth}
\centering\includegraphics[width=1.0\linewidth]{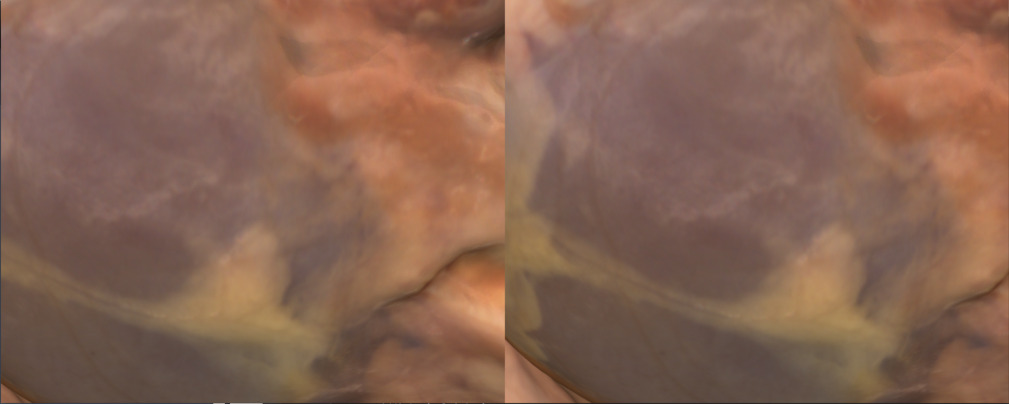}
\subcaption{}
\label{fig:SmoothRGBStereoPair}
\end{subfigure}
\end{tabular}
&
\begin{subfigure}{0.234\textwidth}
\centering\includegraphics[width=1.0\linewidth]{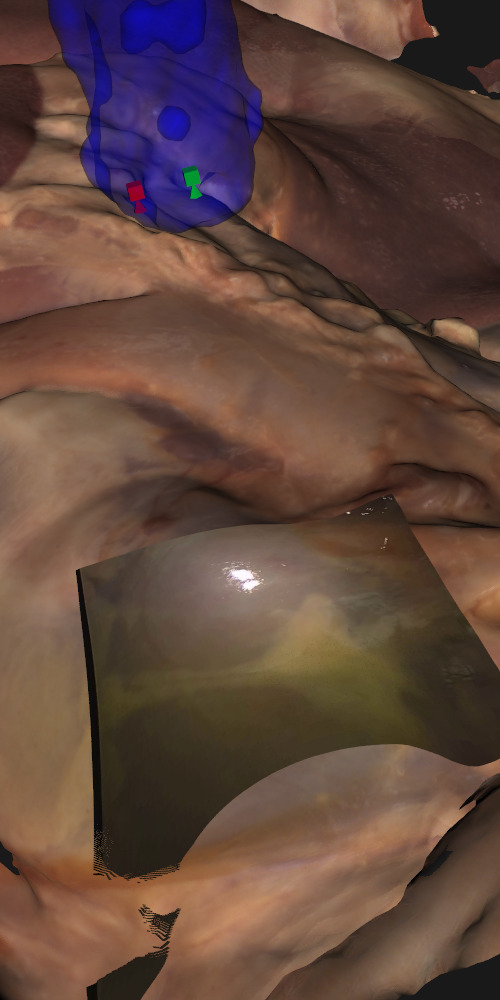}
\subcaption{}
\label{fig:SmoothRGBrendering}
\end{subfigure}
\end{tabular}
}
\caption{Manual alignment for a smooth surface where there is little geometrical variation to register on~\subref{fig:SmoothStereoPair}. It is possible to use the Creaform scanner RGB surface. This provides surface texture and can be readily aligned to features in the image~\subref{fig:SmoothRGBStereoPair}. Overlay of the endoscope image projected onto the RGB surface is also shown~\subref{fig:SmoothRGBrendering}. {\it In all cases in the paper the left image of a stereo pair is shown on the right to facilitate cross-eyed fusion}} 
\label{fig:ManualRGBAlignment}       
\end{figure*}

To achieve an accurate alignment the human eye is a very useful tool. We are able to fuse even the difficult stereo images from surgical scenes and can accurately assess depth. To make use of this human ability, we devised an interactive application that overlays the anatomical surface from the CT scan onto the stereo endoscope view. Some example renderings are shown in Fig.~\ref{fig:ManualAlignment}. The surface can be turned on and off, faded in and out, rendered solid textured or as lines or points. Rendering and interaction use the Visualization Toolkit (Python VTK version 8.1.2) from ~\cite{schroeder2004visualization}. The VTK cameras corresponding to the left and right views are adjusted to match the OpenCV stereo rectification using elements from the SciKit-Surgery library from \cite{thompson2020scikit} (scikit-surgeryvtk version 0.18.1). The underlying rectified left and right images are shown as a background to the rendering. The surface can also be moved, but only the three angles of the endoscope rotation about the left camera can be adjusted. The endoscope position is constrained. This ensures that the perspective from the endoscope is maintained, ensuring that occlusions and the distance to the anatomical surface should be correct.

The X and Y rotations about the endoscope camera are actually very similar to translation in Y and X for small angles. The Z rotation orients the endoscope about its axis, which approximates to a 2D rotation of the image. A small translation along the endoscope axis also allowed to account for the true position of the effective pinhole within the endoscope shaft. The effect is similar to an overall scaling of the image. The left and right images are swapped for display to facilitate cross-eyed fusion of the stereo pair. Coarse and fine adjustments are made until the user is happy with the registration. This is not an easy process, requiring some concentration and skill, but usually takes only a few minutes for each image set (see Fig.~\ref{fig:ManualAlignment}).

\subsubsection{Assessment of manual registration using by repeat alignments}
\begin{figure*}[!t]
\centering
\resizebox{1.0\textwidth}{!}{%
\includegraphics[width=1.0\linewidth]{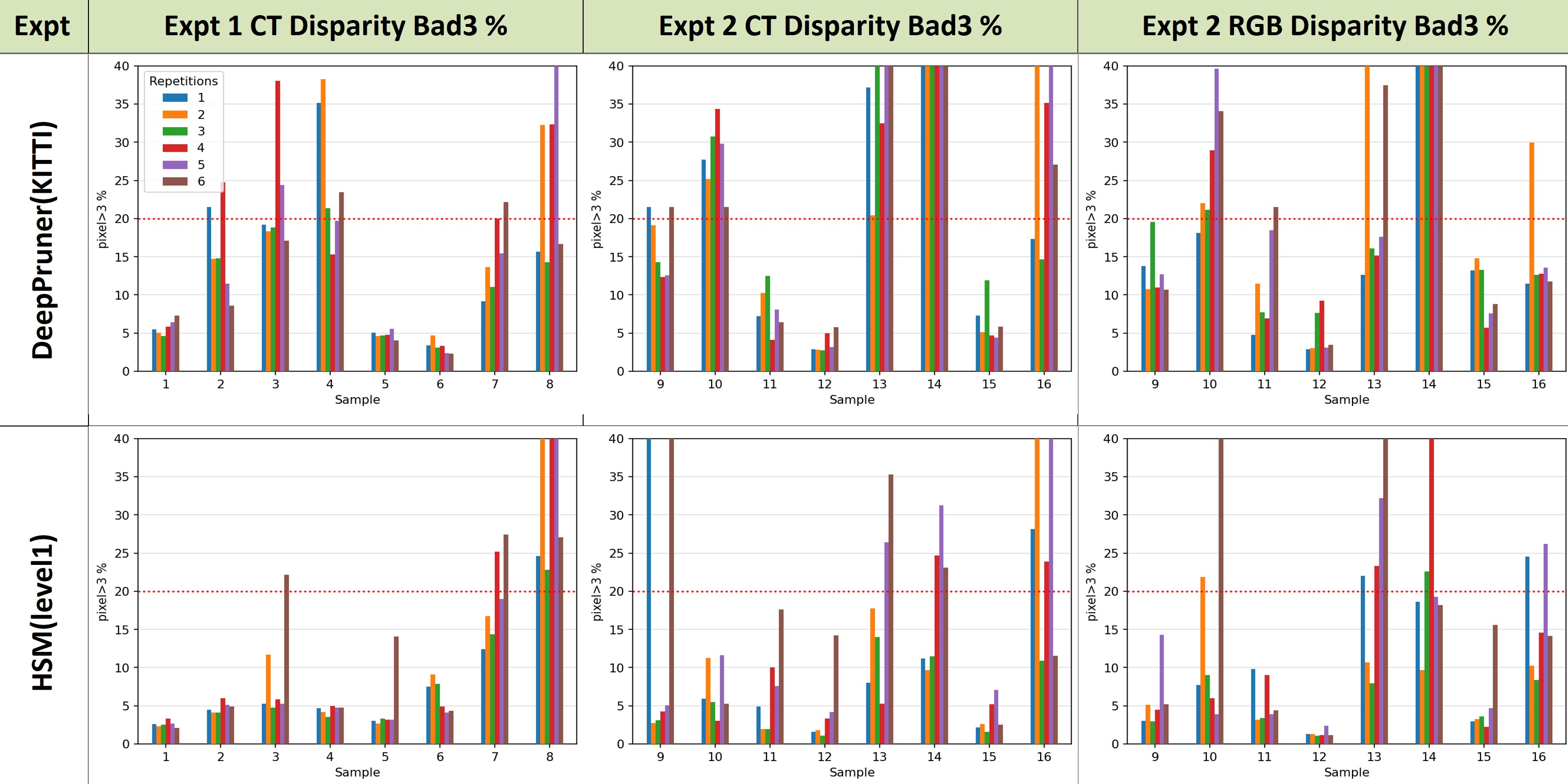}
}
\caption{Bad3 error metric (percentage of disparities $>$3 pixels error) for multiple manual alignments. Results that have $>$20\% Bad3 for both HSM (level 1) and DeepPruner are considered as outliers due to human error. Outliers were not taken into consideration when computing the final dataset as the average of the multiple manual alignments. }
\label{fig:ManualAccuracyEstimate}
\end{figure*}

To assess the accuracy of manual alignment we performed repeat experiments, two people performed the manual alignment 3 times each for every image pair. Based on the evaluation study (\ref{sec:eval}) we chose the he two best performing networks trained in generic data -- HSM (Level 1) and DeepPruner -- to provide independent estimates of the disparity. All error metrics were calculated and the Bad3 results from each alignment can be seen in Fig.~\ref{fig:ManualAccuracyEstimate}. 

It is clear from these results that each network performs better on some images. DeepPruner is better for images 006, 007 and 008, whereas results are similar or HSM (level1) is better for all others. To reduce the effect of human error, we consider manual alignments that have Bad3 $\>$ 20\% compared to both networks as outliers. These are removed before averaging all remaining inlier alignments.

The process of averaging deserves some attention. The $i^{th}$ manual alignment results in a rigid transformation from CT to the left endoscope camera which can be represented by a rotation, $R_i$  and translation $T_i$. We want to find the mean, $\bar{R}, \bar{T}$, of $\{R_i, T_i\}$.

Averaging a set of rotations is a well studied problem and the eigen analysis solution from \cite{markley2007averaging} that is freely available as a NASA report\footnote{\url{https://ntrs.nasa.gov/citations/20070017872}} is widely accepted as an optimal solution. We compute this average using the quaternion representation and provide a mean rotation matrix, $\bar{R}$.

It is sometimes suggested that averaging the translations is trivial or can be achieved by simply taking the mean of $\{T_i\}$, but this is not the case. This would provide the average translation of the CT origin. Our anatomical surface comes from a CT scan where both anatomy and endoscope must be visible, which means that both are pushed towards the periphery of the scan and are some distance from the origin. Using the mean of $\{T_i\}$ leads to a transformed surface that is not centrally placed relative to the manually aligned surfaces.

To obtain a more suitable average translation, we require a more relevant center of rotation in the CT space, $C_{ct}$. In most graphical applications this could be the centroid of an object. In our case a suitable choice would be a point on the viewed CT surface near the centre of endoscope image. For each image pair such a point was identified on the anatomical CT surface with approximately average depth and near to the centre of the endoscope view.

Having chosen $C_{ct}$, calculation proceeds as follows:

\begin{flalign*}
    && \bar{y_i} &= \frac{1}{n} \sum_i{y_i}&\\
\text{where} && y_i &= R_i C_{ct} + T_i&\\
\text{then} && \bar{T} &= \bar{y_i} - \bar{R} C_{ct}
\end{flalign*}

\begin{figure*}[!t]
\centering
\resizebox{1.0\linewidth}{!}{%
\begin{tabular}{ccc}
\includegraphics[height=1.0\linewidth]{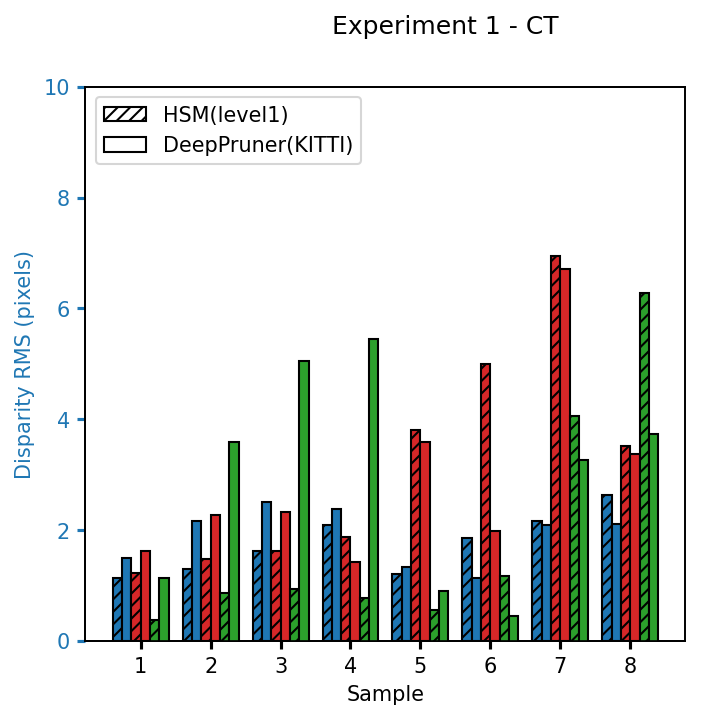} &
\includegraphics[height=1.0\linewidth]{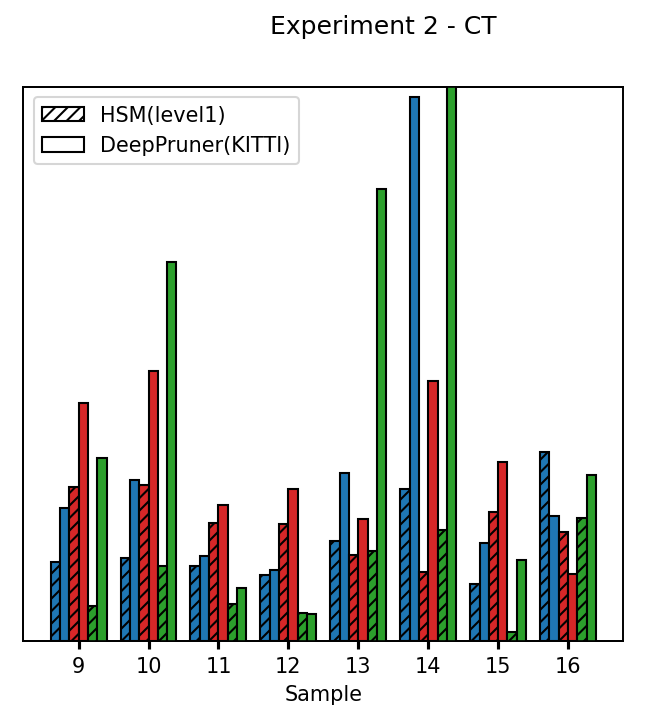} &
\includegraphics[height=1.0\linewidth]{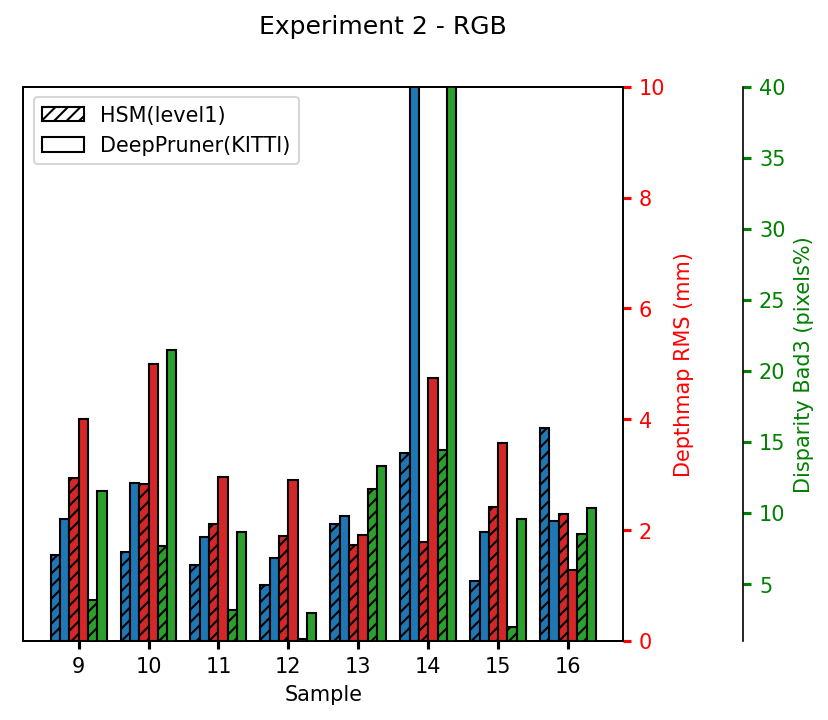}\\
\end{tabular}
}
\caption{Error metrics for the average manual registration of each image. Some very high errors can be attributed to failure of the networks on these difficult images, but at least one network succeeds for each sample. The Disparity has a consistent RMS error of around 2 pixels or less compared to the best performing network. Depth error is also around 2mm in most cases. Sample 7 has higher depth error due to the greater absolute depth (see figure~\ref{fig:disparity_depth_ranges}) which reduces the effect of disparity. The difficult smooth specular images (14 and 16) have slightly higher error, which is likely to be at least in part due to errors from the networks rather than the reference.}
\label{fig:ManualAccuracyDisparityRMSE}
\end{figure*}

With $\bar{R}, \bar{T}$ calculated in this way we achieve a mean rigid transformation that provides a natural average of the manual inputs. Figure~\ref{fig:ManualAccuracyDisparityRMSE} shows the disparity This mean is used in all subsequent calculations and to provide the depth and disparity maps for the released version of SERV-CT.

\subsubsection{Calculation of depth maps, disparity maps and regions of correspondence}
\label{sec:calc_depth_disp}
\begin{figure*}[!t]
\centering
\resizebox{0.9\textwidth}{!}{%
\includegraphics{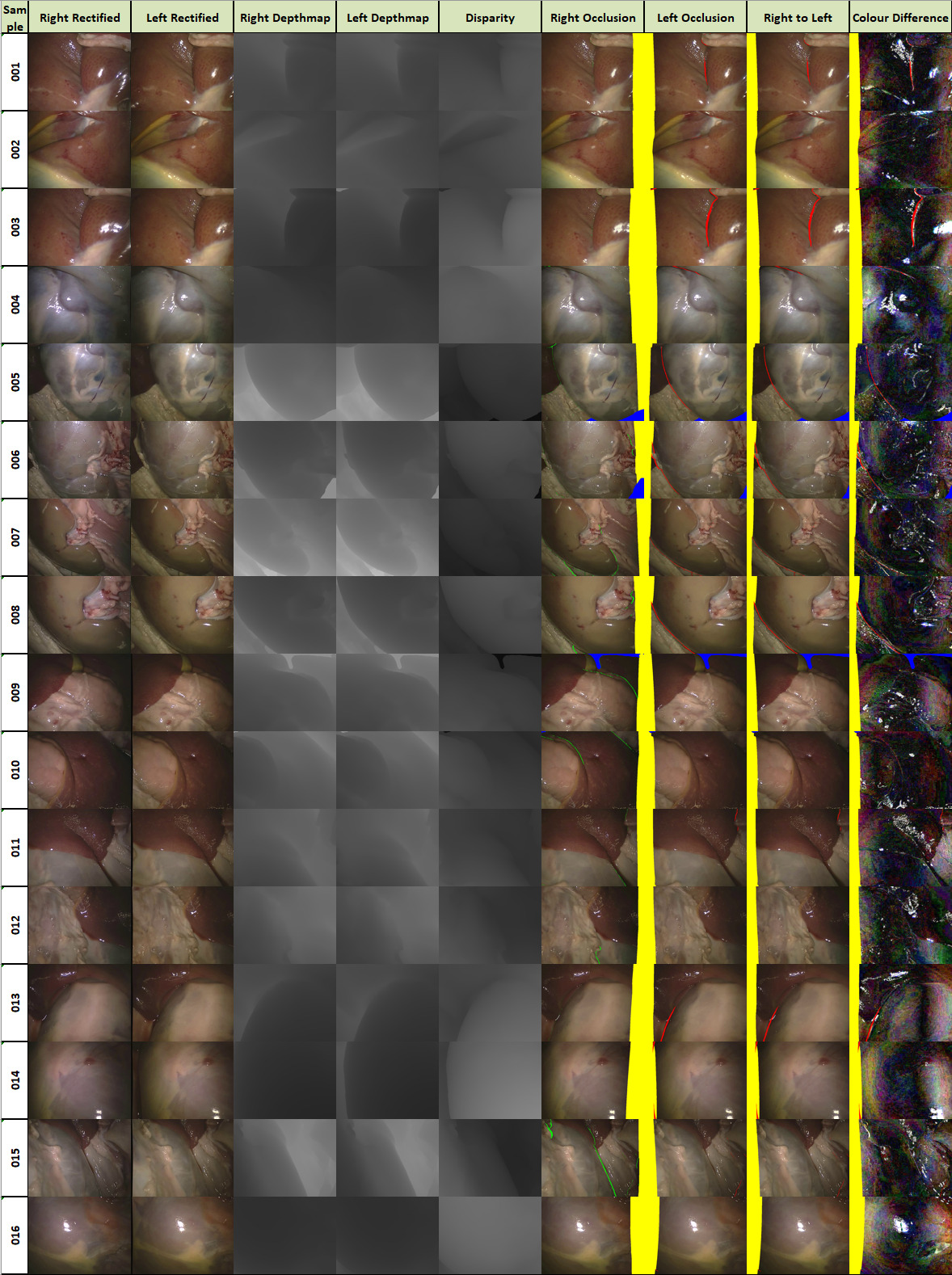}
}
\caption{The images that comprise the SERV-CT testing set. Depth maps are constructed using the OpenGL Z-buffer. Disparity comes from the depth and the stereo rectified $Q$ matrix. Occluded regions are those that have different depths for left and right once correspondence is established using the disparity (non overlap is shown in yellow, occlusion is in red and green, for right and left images respectively, and areas outside the surface model are in blue). Resampled right-to-left and amplified colour difference images are also shown.}
\label{fig:OcclusionsOverlap}
\end{figure*}

Once the CT scan is aligned to the stereo endoscopic view we can produce depth maps and subsequently disparity images. The OpenGL Z-buffer provides a depth value for every pixel. For perspective projection the normalised buffer value, $Z_{gl}$, has a non-linear relationship to actual depth that can be readily calculated from the near and far clipping planes, $Z_{near}$ and $Z_{far}$:
\begin{equation*}
    Z_{world} = {\frac{-2 Z_{near} Z_{far}} {2 (Z_{gl}-0.5) (Z_{far} - Z_{near}) - Z_{near} - Z_{far}}}
\end{equation*}

The $Q$ matrix equation (Eqn.~\ref{eqn:Q_matrix}) can also be easily inverted to provide disparity from depth.
\begin{equation*}
    \delta = (T_x f / Z_{world}) - (C_x^1 - C_x^2)
\end{equation*}
This gives a disparity for every pixel, which is the x displacement between the corresponding point in the left and right images.

A depth map is produced for both the left and right images. Occlusions can be calculated by looking at the depth for a pixel in the left image and its corresponding pixel using the disparity value in the right image. In a rectified setup, the Z coordinate of a point should be the same in the left and right camera coordinates. Any points that do not have the same Z value within an error margin are considered occluded regions visible in the left image but not in the right image. We calculate occlusions for both the left and right images for completeness. A depth equivalent to the back clipping plane corresponds to a point not visible in the CT scan. The resulting images can be seen in Fig.~\ref{fig:OcclusionsOverlap}. The useful pixels for stereo can be seen and cover the majority of the image.

\subsection{Distribution format}
Since there are a number of different packages, formats and coordinate systems for stereo calibration, we distribute the dataset in a much simplified form. We provide only the rectified left and right images with corresponding left and right depth images and disparity from left to right. In addition, we provide a combined mask image for both cameras that shows regions of non-overlap, areas not covered by the 3D model and also identifies occluded pixels which can be seen in one view but not the other. Stereo rectification parameters are provided in a single JSON file containing the $P_1$, $P_2$ and $Q$ matrices for each image. This format can be directly used by any algorithm, neural network or otherwise, that produces disparities from rectified images to compare their output to the reference. We feel that this format significantly simplifies the process and will enable rapid use of the dataset. The original images, calibration of distortion and stereo correspondence, segmented CT scans, surfaces and scripts for registration are also made available in a separate archive.

\section{Evaluation study}
\label{sec:eval}
We use SERV-CT to evaluate the performance of different stereo algorithms. To evaluate disparity outputs we use the bad3 \% error, which is the proportion of the image with greater than 3 pixels disparity error, and root mean square disparity error (RMSE) since these are the popular metrics in stereo evaluation platforms (see \cite{scharstein2014high}).

\subsection{Investigated stereo algorithms}
\label{sec:algorithms}
We primarily focus on the performance of real time deep neural stereo algorithms as these perform best in different challenges. The models were selected based on their error and inference time, as reported on KITTI-15 leader-board. We chose only methods that are publicly available and can provide inference at more than 10 frames per second. We also included PSMNET despite its slower performance since this was the winning technique of a grand challenge in endoscopic video. Github repositories containing both implemented networks and pretrained models used in this study, are provided in the footnote.
Inference runtime performance and GPU memory consumption for each network on a single NVIDIA Tesla V100 can be found in table~\ref{tab:net_performance}. We measure inference time excluding loading time from a storage device to CPU RAM but including prepossessing.

Since a sufficiently large, high quality dataset for training stereo networks for laparoscopy does not yet exist, we choose to use only use networks whose pre-trained weights in a large stereo dataset are available online. This will also allow us to investigate how these networks adapt to a completely different domain from the their training sample and how they compare with the best methods for surgical stereo. All the networks we are testing are regression end-to-end architectures, meaning that they allow gradients to flow freely from the output to their inputs and also produce subpixel disparities. We test some of the fastest deep stereo architectures as inference time performance is an important aspect of any surgical reconstruction system. We also include in our comparison a traditional vision method devised specifically for endoscopic surgical video, the quasi-dense stereo (Stereo-UCL) algorithm by \cite{stoyanov2010real}, which was used in the the TMI study by  \cite{maier2014comparative} and its implementation is available online and as part of OpenCV contrib since version 4.1.

\begin{table}[]
\caption{\label{tab:net_performance} Neural Network performance}
\resizebox{1\linewidth}{!}{%
\begin{tabular}{lcc}
\toprule
{Model} & {GPU Memory (MB)} & {Frames/sec.} \\
\midrule
PSMNet         & 4345                     & 3  \\
DeepPruner     & 1311                     & 16 \\
HSM(level1)    & 1273                     & 20 \\
HSM(level2)    & 1209                     & 28 \\
HSM(level3)    & 1187                     & 32 \\
HAPNet         & 1139                     & 20 \\
DispnetC       & 2311                     & 20 \\
MADNet         & \textbf{1043}            &  \textbf{35}\\
\bottomrule
\end{tabular}
}
\end{table}

\subsubsection{DispNetC}
Along with the introduction of large synthetic stereo dataset for training deep neural networks, \cite{mayer2016large} introduced the first end-to-end stereo network architectures, which achieved similar results the best contemporary methods while being orders of magnitude faster. We will focus on DispNetC\footnote{\url{https://github.com/CVLAB-Unibo/Real-time-self-adaptive-deep-stereo}} which  consists of a feature extraction module, a correlation layer and a encoder decoder part with skip connections which aims to refine the cost volume and compute disparities. The feature extraction part of this network downsamples and extracts unary features for each image separately. Those features get correlated together in the horizontal dimension, building a cost volume, which in turn gets further refined. This last refinement and disparity computation sub-module, follows an 2D encoder-decoder architecture, with skip connections, which is in place to allow matching for large disparities and provides subpixel accuracy.

\subsubsection{PSMNet}
Influenced by work in semantic segmentation literature, \cite{chang2018pyramid} introduced PSMNet\footnote{\url{https://github.com/JiaRenChang/PSMNet}}, achieving then leading performance on the KITTI leaderbord. Although it cannot be considered real time, its novelty lies on the incorporation of spatial pyramidal pooling (SPP) module. This architecture enables such networks to extract unary features that take into account global context information, something that is crucial in surgical stereo applications due to homogeneous surface texture or the presence of specular highlights. The SPP module achieves this by extracting features at different scales and later concatenates them before forming a 4D feature volume. This 4D feature volume gets refined by a stacked hourglass architecture which further improves disparity estimation.

\subsubsection{HSM}
\cite{yang2019hsm}, in an effort to develop a network that can infer depth fast in close range, to be used in autonomous vehicles platforms, developed the hierarchical deep stereo matching (HSM) network\footnote{\url{https://github.com/gengshan-y/high-res-stereo}}. This allows fast and memory efficient disparity computation allowing it to process high resolution images. In the feature extraction process, the down-sampled feature images form feature volumes, each, corresponding to a different depth scale. Feature volumes of coarser scales get refined, upsampled and concatenated to the one of the next finer scale, hierarchically refining the disparity estimation. This enables the network to make fast queries from intermediate scales in the expense of depth resolution.

\subsubsection{DeepPruner}
DeepPruner\footnote{\url{https://github.com/uber-research/DeepPruner}} is another fast and memory efficient deep stereo architecture from~\cite{Duggal2019ICCV}. Based on the PatchMatch algorithm from \cite{barnes2009patchmatch}, they created a fully differentiable version of it making ideal to incorporate it in a end-to-end neural network. The PatchMatch module prunes the disparity search space, enabling the network to search in a smaller disparity range, which in turn reduces the memory consumption, as well as the time to build and process this feature volume. The feature volume gets processed by a refinement network to increase matching performance. The differentiable PatchMatch module, though it does not contain any learnable parameters, enables gradient flow, facilitating end-to-end training. In our test we use the fast configuration of this method as described in the original paper.

\subsubsection{MADNet}
To tackle to domain shift problem most deep learning architecture experience, \cite{tonioni2019real} introduced modularly adaptive network, MADNet\footnote{\url{https://github.com/CVLAB-Unibo/Real-time-self-adaptive-deep-stereo}}, which can be used with the modular adaptation (MAD) algorithm enabling the network to adapt in a different target domain from the one that it's trained on. The architecture is one of the fastest in the literature and the online modular adaptation scheme is efficient enough to run in real time. MADNet is based on a hierarchical pyramid and cost volume architecture, which enables it to employ the adaptation scheme at inference time, without reducing real time performance significantly. In our experiments we do not use the MAD because the number of available samples are limited.

\subsubsection{HAPNet}
3D convolutions and manual feature alignment are the two least efficient operations deep stereo neural networks perform. Working towards a real time stereo matching network, \cite{brandao2020hapnet} introduced HAPNet\footnote{\url{https://github.com/patrickrbrandao/HAPNet-Hierarchically-aggregated-pyramid-network-for-real-time-stereo-matching}}, an architecture that extract features in different scales, concatenates them and find correspondences using a 2D hourglass encoder decoder block. The disparity estimation process is done in a hierarchical fashion, where low resolution feature maps are used to find low resolution disparities. Those low resolution disparities get up-convolved and concatenated with the features of the next scale to hierarchically refine and regress the final disparity. The 2D hourglass encodes correspondences from the concatenated features, effectively enlarging the receptive field of the network.

\subsubsection{Stereo-UCL}
To estimate depth in surgical environments that specular reflections and homogeneous texture make most stereo algorithms fail, \cite{stoyanov2010real} developed an algorithms we will refer to as Stereo-UCL\footnote{Available in OpenCV version 4.1 and above} to be compatible with the TMI evaluation study. The algorithms produces semi-dense disparities based on a best first region growing scheme. It is the only algorithm out of the methods of this study that can robustly estimate matches between unrectified surgical stereo pairs. In an initial step, the method finds sparse features in the left frame and then it uses optical flow to match with pixels in the right view. Those pixels are used as inputs seeds in a region growing algorithm that propagates disparity information from known disparities to adjacent pixels.

\subsection{Error Metrics}
We use error metrics based on those generally reported in the literature. We measure distance in 3D as the difference between reference depthmaps and the depth computed based on disparity outputs of the investigated methods using the $Q$ matrix. In the 2D case we chose to use the both Bad3 \% error, which is the percentage of disparity pixels that diverge from the reference more than 3 pixels, and 2D RMS error, taking the mean across all samples.

\subsection{Evaluation details}
For each deep learning model and method we run inference across all samples, using models trained on mainstream computer vision datasets (\cite{Menze2015CVPR,mayer2016large}). All deep learning based methods are configured to search for matches up to at least 192 pixels, and results are stored as 16-bit PNG images, normalized appropriately to encode subpixel information. We use the depthmap supplied as part of SERV-CT as previously described and compare this to a triangulated depthmap from the network output. For disparity evaluation, since we have disparities from both SERV-CT and from the stereo algorithms, we can directly measure the error without further processing. When evaluating the results, both for 2D and 3D, we include separate results for occluded and non-occluded pixels. We separate the results for the first {\it ex vivo} sample (Expt. 1 - CT) and the second sample where we have both the CT surface (Expt. 2 - CT) and the RGB surface from the Creaform scanner (Expt. 2 - RGB).

\section{Results}
\label{sec:results}


\begin{figure*}
\resizebox{1\textwidth}{!}{%
\begin{tabular}{cc}
\includegraphics{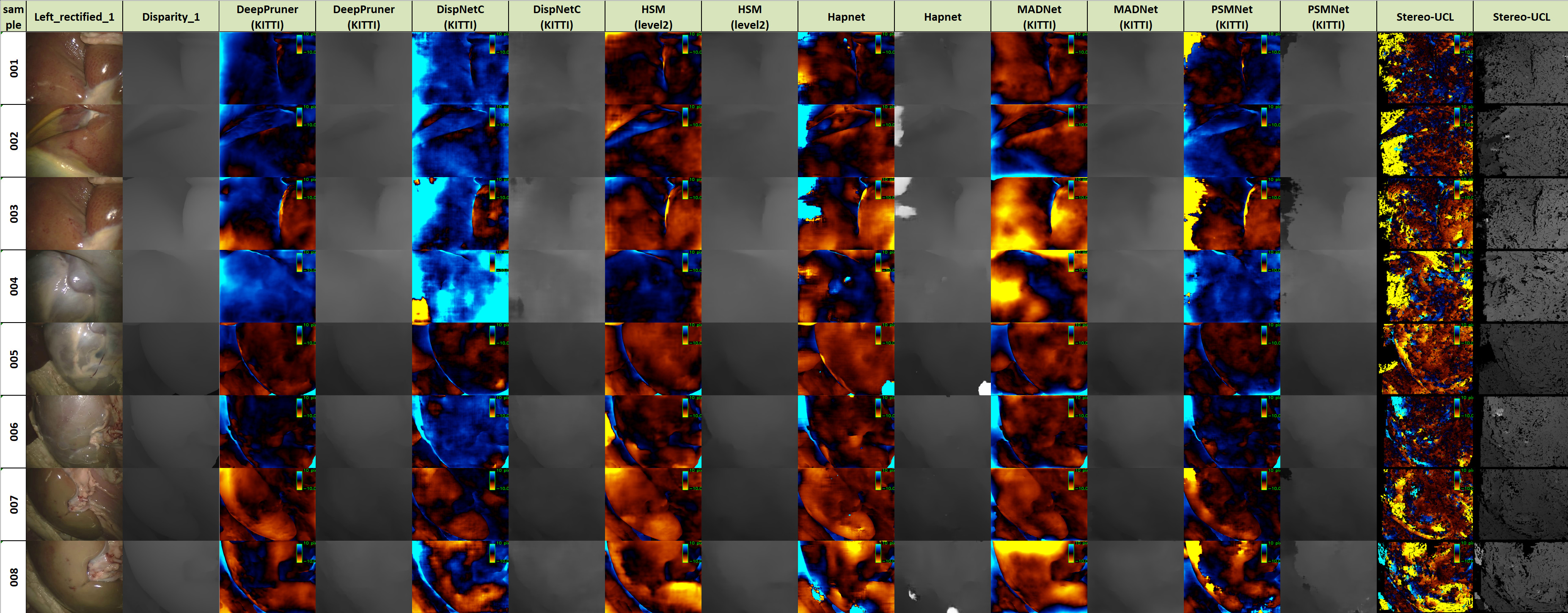} &
\raisebox{\linewidth}{
\begin{tabular}{c}
{\Huge 10} \\
\includegraphics[width=0.1\linewidth]{ColourScale.png} \\
{\Huge -10} \\
\end{tabular}
}
\end{tabular}
}
\caption{Signed disparity error in pixels of each algorithm compared to the CT reference for Expt. 1 (hotcold colormap from \href{https://github.com/endolith/bipolar-colormap}{endolith})}
\label{fig:ResultsImageTable_Expt1_CT}
\end{figure*}

\begin{table*}[!t]
\caption{\label{tab:expt1_CT} Expt. 1 - CT results}
\resizebox{\linewidth}{!}{%
\begin{tabular}{lrrrrrr}
\toprule
{} & \multicolumn{2}{c}{Mean Bad3 Error} & \multicolumn{2}{c}{Mean RMSE} & \multicolumn{2}{c}{Mean RMSE} \\
{} & \multicolumn{2}{c}{\%} & \multicolumn{2}{c}{3D Distance (mm)} & \multicolumn{2}{c}{Disparity (pixels)} \\
\hspace{2cm} Occlusions: &             not included &   included &                    not included &    included &                  not included &   included \\
Method                &                 &       &                        &        &                      &       \\
\midrule

DeepPruner(KITTI)     & 12.50  (±7.44) & 17.18  (±8.18) &    \textbf{2.91}    (±1.71) &   \textbf{3.77}    (±1.65) &  1.91  (±0.51) &  2.47  (±0.60) \\
DeepPruner(SceneFlow) & 53.63 (±20.39) & 56.27 (±19.70) &   13.01    (±8.60) &  17.09    (±8.79) & 18.29 (±14.44) & 24.35 (±15.69) \\
DispNetC(KITTI)       & 40.09 (±26.41) & 42.79 (±27.32) &    4.58    (±0.76) &   5.66    (±0.95) &  4.24  (±2.70) &  5.20  (±3.29) \\
DispNetC(SceneFlow)   & 26.78 (±17.77) & 31.00 (±19.13) &    4.66    (±1.52) &   6.22    (±2.14) &  4.23  (±2.59) &  5.75  (±4.20) \\
HSM(level1)           &  \textbf{8.34}  (±8.31) & \textbf{10.84}  (±8.93) &    3.18    (±2.03) &   4.43    (±2.91) &  \textbf{1.75}  (±0.53) &  \textbf{2.19}  (±0.69) \\
HSM(level2)           & 12.28 (±10.20) & 14.76 (±10.21) &    3.53    (±2.16) &   4.64    (±2.65) &  2.13  (±0.69) &  2.50  (±0.74) \\
HSM(level3)           & 63.90 (±12.27) & 63.00 (±11.93) &   10.09    (±7.07) &  10.62    (±7.16) &  5.42  (±1.29) &  5.50  (±1.37) \\
Hapnet                & 17.85 (±13.31) & 21.35 (±13.11) &    6.01    (±4.07) &   8.27    (±3.94) &  7.41  (±7.40) & 12.47 (±11.38) \\
MADNet(KITTI)         & 26.58 (±18.11) & 30.09 (±18.28) &    4.23    (±1.42) &   5.05    (±1.50) &  3.44  (±1.64) &  3.83  (±1.62) \\
MADNet(SceneFlow)     & 34.01 (±16.31) & 39.52 (±15.49) &   13.22   (±13.31) &  16.48   (±13.69) & 15.75 (±14.89) & 19.78 (±14.95) \\
PSMNet(KITTI)         & 12.16  (±7.12) & 17.37  (±8.15) &    9.53    (±9.40) &  18.15   (±21.23) &  3.48  (±2.21) &  5.59  (±4.32) \\
PSMNet(SceneFlow)     & 98.31  (±1.23) & 98.11  (±1.11) &   19.63    (±6.33) &  22.52    (±7.13) & 24.60  (±6.22) & 31.61  (±9.23) \\
Stereo-UCL            & 33.24 (±10.09) & 34.43 (±10.18) &   26.40   (±19.55) &  36.46   (±32.03) &  9.24  (±2.96) & 10.89  (±4.29) \\

\bottomrule
\end{tabular}
}
\end{table*}

\begin{figure*}
\resizebox{1\textwidth}{!}{%
\begin{tabular}{cc}
\includegraphics{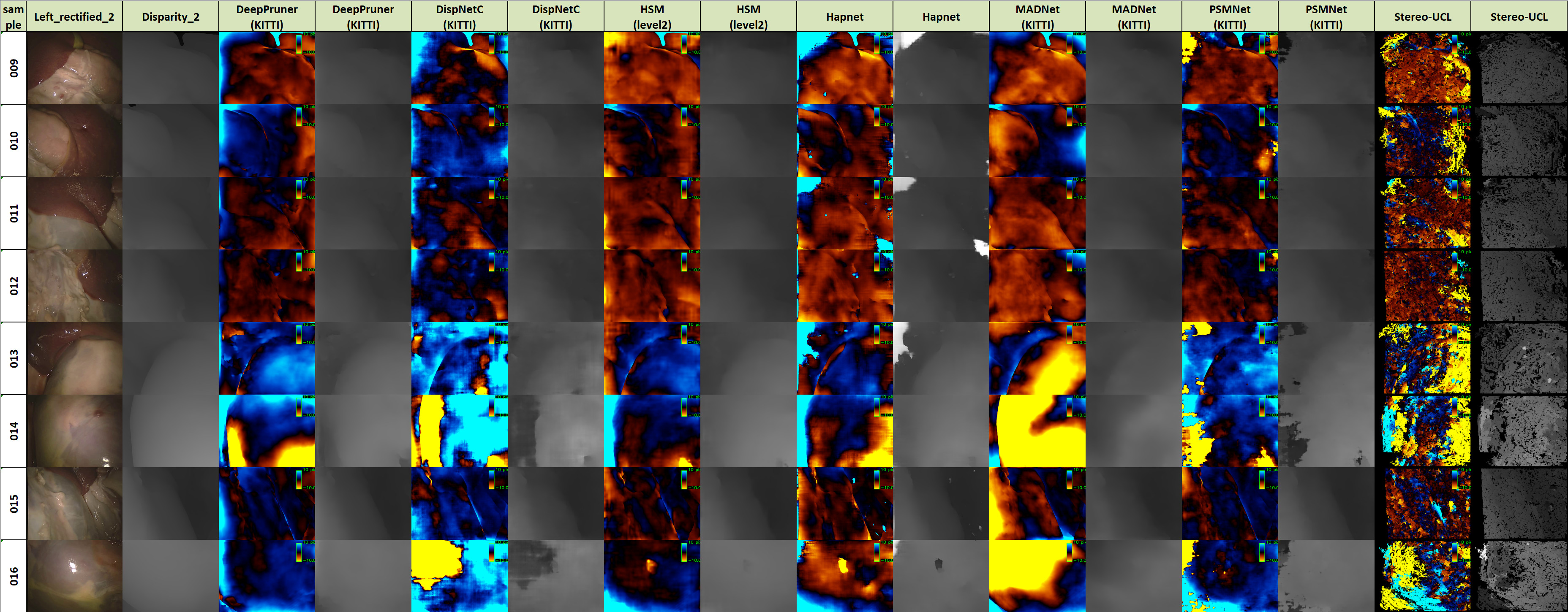} &
\raisebox{\linewidth}{
\begin{tabular}{c}
{\Huge 10} \\
\includegraphics[width=0.1\linewidth]{ColourScale.png} \\
{\Huge -10} \\
\end{tabular}
}
\end{tabular}
}
\caption{Signed disparity error in pixels of each algorithm compared to the CT reference for Expt. 2 (hotcold colormap from \href{https://github.com/endolith/bipolar-colormap}{endolith})}
\label{fig:ResultsImageTable_Expt2_CT}
\end{figure*}

\begin{table*}[!t]
\caption{\label{tab:expt2_CT} Expt. 2 - CT results}
\resizebox{\linewidth}{!}{%
\centering
\begin{tabular}{lrrrrrr}
\toprule
{} & \multicolumn{2}{c}{Mean Bad3 Error} & \multicolumn{2}{c}{Mean RMSE} & \multicolumn{2}{c}{Mean RMSE} \\
{} & \multicolumn{2}{c}{\%} & \multicolumn{2}{c}{3D Distance (mm)} & \multicolumn{2}{c}{Disparity (pixels)} \\
\hspace{2cm} Occlusions: &             not included &   included &                    not included &    included &                  not included &   included \\
Method                &                 &       &                        &        &                      &       \\
\midrule

DeepPruner(KITTI)     & 19.13 (±16.95) & 24.58 (±15.75) &   3.21   (±1.31) &   3.82  (±1.65) &   3.12  (±2.77) &  4.03  (±2.71) \\
DeepPruner(SceneFlow) & 87.73 (±15.79) & 88.99 (±14.13) &  29.77  (±11.76) &  33.14 (±12.52) &  55.27 (±16.00) & 68.05 (±13.15) \\
DispNetC(KITTI)       & 47.87 (±27.58) & 50.72 (±25.40) &   7.07   (±4.56) &   7.49  (±4.70) &   7.53  (±7.33) &  8.12  (±7.20) \\
DispNetC(SceneFlow)   & 47.60 (±32.15) & 50.25 (±31.02) &   7.68   (±3.68) &   8.77  (±4.67) &  15.21 (±17.76) & 17.09 (±20.14) \\
HSM(level1)           &  \textbf{5.46}  (±2.96) &  \textbf{9.22}  (±5.20) &   \textbf{2.12}   (±0.54) &   \textbf{2.98}  (±1.29) &   \textbf{1.81}  (±0.83) &  \textbf{2.73}  (±2.07) \\
HSM(level2)           &  9.73  (±3.96) & 13.68  (±5.72) &   2.78   (±1.17) &   3.74  (±1.24) &   2.07  (±0.63) &  3.08  (±1.92) \\
HSM(level3)           & 46.95 (±11.61) & 49.77  (±9.73) &   6.47   (±3.68) &   7.43  (±3.09) &   4.57  (±1.00) &  5.68  (±2.35) \\
Hapnet                & 23.64 (±17.61) & 27.71 (±16.56) &   8.45   (±4.99) &  10.33  (±5.24) &  13.09  (±8.88) & 18.82 (±12.96) \\
MADNet(KITTI)         & 38.24 (±30.26) & 40.39 (±27.63) &   6.31   (±3.13) &   6.72  (±3.89) &   7.21  (±7.02) &  7.68  (±7.25) \\
MADNet(SceneFlow)     & 69.49 (±19.68) & 71.61 (±18.34) &  30.17  (±15.19) &  31.24 (±12.45) &  46.47 (±21.83) & 50.59 (±18.03) \\
PSMNet(KITTI)         & 15.41 (±13.02) & 19.58 (±15.52) &   9.19   (±7.67) &  12.94 (±12.58) &   5.09  (±4.99) &  6.65  (±5.85) \\
PSMNet(SceneFlow)     & 96.18  (±1.58) & 96.38  (±1.30) &  23.78   (±9.26) &  25.17  (±8.76) &  35.85 (±17.75) & 40.98 (±18.94) \\
Stereo-UCL            & 42.74 (±22.25) & 43.05 (±22.44) &  35.19  (±37.62) &  35.63 (±37.49) &  12.45  (±8.01) & 13.10  (±8.82) \\

\bottomrule
\end{tabular}
}
\end{table*}


\begin{figure*}
\resizebox{1\textwidth}{!}{%
\begin{tabular}{cc}
\includegraphics{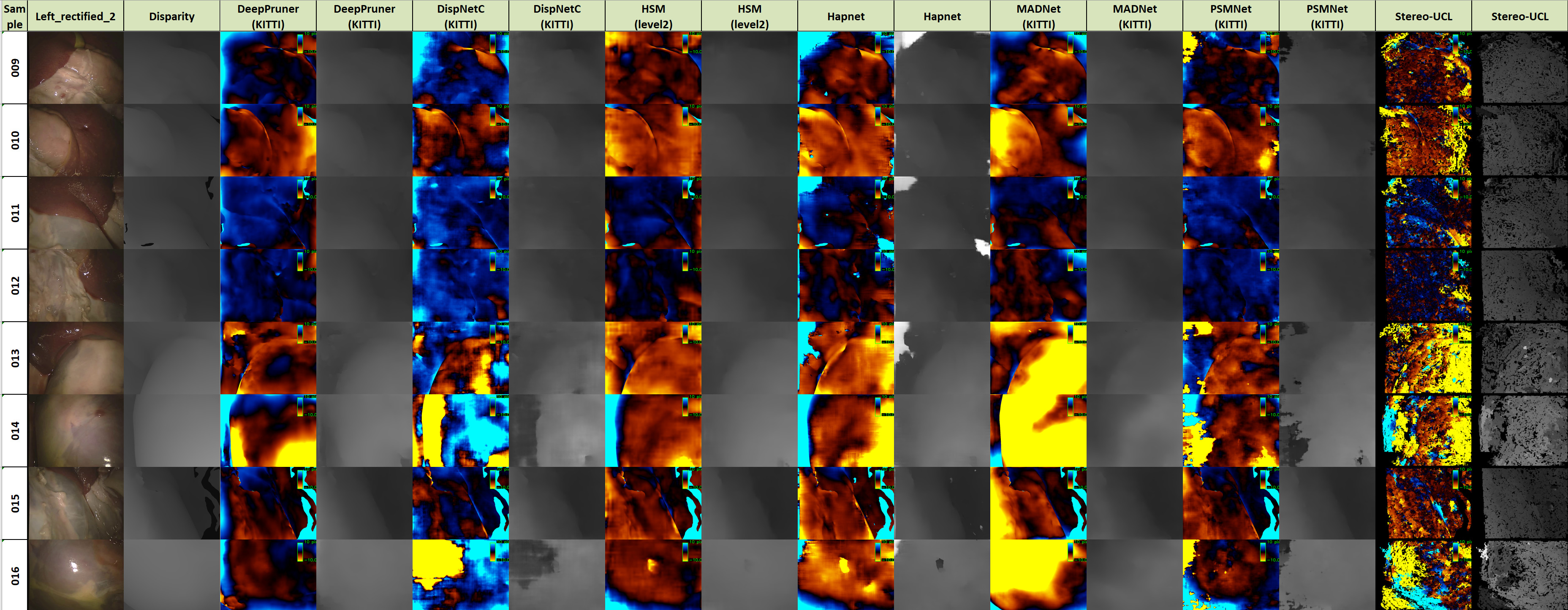} &
\raisebox{\linewidth}{
\begin{tabular}{c}
{\Huge 10} \\
\includegraphics[width=0.1\linewidth]{ColourScale.png} \\
{\Huge -10} \\
\end{tabular}
}
\end{tabular}
}
\caption{Signed disparity error in pixels of each algorithm compared to the Creaform RGB  reference for Expt. 2 (hotcold colormap from \href{https://github.com/endolith/bipolar-colormap}{endolith})}
\label{fig:ResultsImageTable_Expt2_Creaform}
\end{figure*}

\begin{table*}[!t]
\caption{\label{tab:expt2_Creaform} Expt. 2 - Creaform results}
\resizebox{\linewidth}{!}{%
\centering
\begin{tabular}{lrrrrrr}
\toprule
{} & \multicolumn{2}{c}{Mean Bad3 Error} & \multicolumn{2}{c}{Mean RMSE} & \multicolumn{2}{c}{Mean RMSE} \\
{} & \multicolumn{2}{c}{\%} & \multicolumn{2}{c}{3D Distance(mm)} & \multicolumn{2}{c}{Disparity(pixels)} \\
\hspace{2cm} Occlusions: &             not included &   included &                    not included &    included &                  not included &   included \\
Method                &            &          &       &          &                   &                    \\
\midrule

DeepPruner(KITTI)     &  15.31 (±12.91) & 21.62 (±12.51) &   3.30   (±1.30) &   3.91   (±1.59) &   3.11  (±2.84) &  3.99  (±2.76) \\
DeepPruner(SceneFlow) &  87.06 (±17.17) & 88.44 (±15.26) &  29.36  (±11.75) &  32.86  (±12.55) &  54.62 (±16.35) & 67.77 (±13.13) \\
DispNetC(KITTI)       &  48.24 (±24.92) & 51.05 (±22.92) &   7.19   (±4.43) &   7.60   (±4.60) &   7.46  (±7.15) &  8.08  (±7.13) \\
DispNetC(SceneFlow)   &  47.99 (±30.72) & 50.48 (±29.93) &   7.71   (±3.46) &   8.86   (±4.60) &  15.01 (±17.34) & 16.92 (±19.80) \\
HSM(level1)           &   \textbf{6.58}  (±4.82) & \textbf{10.40}  (±7.05) &   \textbf{2.25}   (±0.45) &   \textbf{3.17}   (±1.54) &   \textbf{2.00}  (±1.06) &  \textbf{2.96}  (±2.38) \\
HSM(level2)           &  10.98  (±5.99) & 15.12  (±8.13) &   2.96   (±1.20) &   3.95   (±1.52) &   2.29  (±0.91) &  3.32  (±2.23) \\
HSM(level3)           &  49.78 (±13.02) & 52.30 (±10.49) &   6.48   (±3.53) &   7.51   (±3.07) &   4.84  (±1.19) &  5.98  (±2.62) \\
Hapnet                &  27.07 (±20.93) & 31.15 (±20.08) &   8.72   (±5.27) &  10.60   (±5.46) &  13.46  (±9.14) & 19.11 (±13.08) \\
MADNet(KITTI)         &  40.28 (±32.16) & 42.27 (±29.22) &   6.49   (±3.25) &   6.92   (±4.00) &   7.55  (±7.28) &  8.06  (±7.53) \\
MADNet(SceneFlow)     &  68.25 (±19.47) & 70.58 (±18.16) &  30.17  (±14.94) &  31.26  (±12.20) &  46.38 (±22.09) & 50.52 (±18.37) \\
PSMNet(KITTI)         &  14.05  (±8.03) & 18.47 (±10.70) &   9.34   (±7.48) &  13.08  (±12.34) &   5.16  (±4.85) &  6.76  (±5.71) \\
PSMNet(SceneFlow)     &  96.26  (±1.22) & 96.47  (±0.96) &  23.85   (±9.67) &  25.28   (±9.16) &  36.11 (±17.95) & 41.21 (±19.10) \\
Stereo-UCL            &  43.72 (±23.50) & 44.02 (±23.64) &  35.03  (±37.87) &  35.48  (±37.72) &  12.68  (±8.35) & 13.34  (±9.14) \\

\bottomrule
\end{tabular}
}
\end{table*}

Results are split into three groups - {\it ex vivo} sample 1 with CT reference (Expt. 1 -- CT), and {\it ex vivo} sample 2 with reference from either the CT scan (Expt. 2 - CT) or Creaform RGB surface registered to CT (Expt. 2 -- RGB).

Numerical results are summarised in tables~\ref{tab:expt1_CT}, \ref{tab:expt2_CT} and \ref{tab:expt2_Creaform}. There is a general trend that deep neural methods trained only synthetic data produce disproportionately high error when compared to versions trained on real data. Networks fine-tuned on real data perform slightly better than the domain specific classical stereo method (Stereo-UCL). HSM and DeepPruner consistently performed the best across all dataset cases. We do not consider networks trained on synthetic data in subsequent analysis.

The greater consistency of the 3D error across our experiments suggests that this may be a more reliable metric. Additionally, 3D error metrics give an idea of the performance in real world distances. However, disparity metrics enable comparison of matching performance on data from potentially very different camera setups.

Figures~\ref{fig:ResultsImageTable_Expt1_CT}, \ref{fig:ResultsImageTable_Expt2_CT} and \ref{fig:ResultsImageTable_Expt2_Creaform} provide error images showing the difference between the reference and the result from each algorithm for every pixel. Error metrics are lower for Expt. 1. Some more challenging images are presented in Expt. 2, particularly samples 013, 014 and 016. These depict smooth surfaces that are either featureless or contain significant specular highlights and clearly present a problem for some of the networks. The difficulty of these images may account for much of the increase in error.


The Bad3 \% error results for our dataset are higher than those generally reported in the computer vision literature. This is mainly because the tested methods are not fine-tuned for this specific dataset. The limited sample size does not facilitate training on this dataset. The challenging images presented to the algorithms and also any inaccuracies in the CT or Creaform reference surfaces will contribute to this error. It is hard to separate algorithm fitting error from reference error, but the spread of performance from all methods suggests that inaccuracies from the reconstruction algorithms may dominate.

The analysis has slightly higher error for the Creaform surfaces for Expt. 2, which may result from the extra registration process from the RGB surface to the CT scan. However, the ability to register using visible surface features that have no corresponding geometric variation is potentially useful and may offer a route to automated alignment.

Overall, we can see that algorithmic performance can be compared using our dataset and that the images included present different challenges to that offered by existing datasets. Although the data has limitations, we believe the value of providing direct disparity maps for evaluation will significantly ease the process of evaluating new algorithms and support the community by establishing benchmarks.

\section{Conclusions}
Standardised datasets have been instrumental in accelerating the development of algorithms for a wide range of problems in computer vision and medical image computing. They not only alleviate the need for data generation but also provide a means of transparently and measurably benchmarking progress. Despite recent progress and emergence of datasets in endoscopy, for example for gastroenterology (the GIANA challenge~\footnote{\url{https://giana.grand-challenge.org/}} and the KVASIR dataset from \cite{pogorelov2017kvasir}), instrument and scene segmentation (\cite{allan20192017, maierhein2020heidelberg, ALHAJJ201924}, and for 3D reconstruction (\cite{maier2014comparative, rau2019implicit, penza2018endoabs}), there is still an unmet need for such high quality stereoscopic datasets for 3D reconstruction in surgical endoscopy, which presents specific challenges due to the lack of clearly identifiable features, highly deformable tissue and the presence specular highlights, smoke, tools and blood.

With this paper, we have developed and reported a validation dataset based on CT images of the endoscope and the viewed anatomical surface. The location of the endoscope constrains the perspective from which the anatomical surface is viewed. Subsequent rotation of the view followed by a small Z translation for scale is established by manual alignment. Constraining the endoscope location in this way ensures that the distance from camera to anatomy comes from the CT dataset. This method of constrained alignment could be used for any 3D modelling system that covers both the endoscope and the viewed anatomy. The dataset and algorithms for processing the raw data and the generated disparity maps are openly available\footnote{\url{https://www.ucl.ac.uk/interventional-surgical-sciences/serv-ct}}. In addition, we report the performance of various real-time computational stereo algorithms that are open source and provide the full parameter settings alongside the trained model weights to allow experiments to be reproduced easily.

SERV-CT adds to existing evaluation sets for stereoscopic surgical endoscopic reconstruction. The validation covers the majority of the image and there is considerable variation of depth in the viewed scene (see Fig.~\ref{fig:disparity_depth_ranges}). The analysis of several stereo reconstruction algorithms has been performed and demonstrates the feasibility of SERV-CT as a validation set, but also highlights challenges. The results suggest that some of the best methods based on real world scenes are promising candidates for surgical stereo-endoscopic reconstruction.

There are limitations to the work. A relatively small number of frames with corresponding reference are provided. The manual alignment relies heavily on operator skill and is not a trivial process. Automating this part of the procedure would be desirable but presents an open registration problem in its own right. There is variation of anatomy and considerable variation of depth in the images presented, but further realism could be provided by the introduction of tools, smoke and blood. The comparatively high bad3 errors could in part come from 3D segmentation, calibration or registration but also reflect the difficult images incorporated into the database. The endoscope images from the original da Vinci are comparatively low contrast and resolution compared to newer endoscopes and there is noticeable colour difference between the eyes.

Datasets are not only important for validation but also for training deep learning models. Many more images would be required, however, for training of neural networks. Possible ways of addressing this are through simulation (\cite{rau2019implicit, pfeiffer2019generating}), but fine tuning may still be required. In our future work plans, we intend to extend this dataset significantly in a variety of ways. Kinematic tracking may extend the CT alignment to multiple frames, significantly increasing the number of frames available from these datasets and also providing a reference for video-based reconstruction and localisation methods such as SLAM. We intend to gather more such datasets under different conditions and will also investigate the use of other devices for measurement of the tissue surface, such as laser range finders and structured light.

Despite the recognised limitations of the SERV-CT dataset, we have established the feasibility of this methodology of reference generation. The method could be applied to any measurement system that can provide the location of both the endoscope and the viewed anatomy in the same coordinate system. It may potentially be a way of approaching the bottlenecks in image-guided surgery through preoperative and intraoperative surface registration. We also hope that this work encourages further development of such reference surgical endoscopic datasets to facilitate research in this important area which may help provide surgical guidance and is likely to be of significant benefit in the development of future robotic surgery.

\section{Acknowledgements}
The work was supported by the Wellcome/EPSRC Centre for Interventional and Surgical Sciences (WEISS) [203145Z/16/Z]; Engineering and Physical Sciences Research Council (EPSRC) [EP/P027938/1, EP/R004080/1, EP/P012841/1]; The Royal Academy of Engineering [CiET1819/2/36].
\typeout{}


\end{document}